\documentclass{article}

 \usepackage[eandd, final]{neurips_2026}
\usepackage[utf8]{inputenc} % allow utf-8 input
\usepackage[T1]{fontenc}    % use 8-bit T1 fonts
\usepackage{hyperref}       % hyperlinks
\usepackage{url}            % simple URL typesetting
\usepackage{booktabs}       % professional-quality tables
\usepackage{amsfonts}       % blackboard math symbols
\usepackage{nicefrac}       % compact symbols for 1/2, etc.
\usepackage{microtype}      % microtypography
\usepackage{xcolor}         % colors
\usepackage{graphicx}

\usepackage{tcolorbox}
\usepackage{fancyvrb}
\usepackage{placeins}
\hypersetup{hidelinks}
\usepackage{hyperref}

\DefineVerbatimEnvironment{gamecodeinner}{Verbatim}{
  fontsize=\footnotesize,
  frame=single,
  framesep=3pt,
  rulecolor=\color{gray!60},
  fillcolor=\color{gray!5}
}

\newcounter{gamecode}

\newenvironment{gamecode}{%
  \gamecodeinner
}{%
  \endgamecodeinner
}

\title{CivBench: A Long-Horizon Benchmark for Tool-Mediated Agents in Civilization VI}

\author{%
    Austin Tudor David Andrews\thanks{Joint first authors; equal contribution.} \\
    University of Oxford
    \And
    Liam Wilkinson\footnotemark[1] \\
    Tony Blair Institute for Global Change
    \And
  Jamie Heagerty \\
  Google DeepMind \\
  \And
  Harry Coppock \\
  UK AI Security Institute \\
  Imperial College London \\
  10 Downing Street \\
  \And
  Jakob Nicolaus Foerster \\
  University of Oxford \\
  \And
  Rui Ponte Costa \\
  University of Oxford \\
}

\hypersetup{
    pdftitle={CivBench: A Long-Horizon Benchmark for Tool-Mediated Agents in Civilization VI},
    pdfauthor={Austin Tudor David Andrews, Liam Wilkinson, Jamie Heagerty, Harry Coppock, Jakob Nicolaus Foerster, Rui Ponte Costa},
    pdfsubject={NeurIPS 2026 Evaluations and Datasets},
    pdfkeywords={CivBench, LLM Agents, Long-Horizon Planning, Model Context Protocol, Benchmark},
    hidelinks
}

\begin{document}
% CivBench paper — auto-generated data macros
% Source: analysis/admissible_games.py + analysis/plots.py + analysis/rag_auto_results.csv
% Regenerate with: .venv/bin/python analysis/plots.py  (also recomputes figures)
% ─────────────────────────────────────────────────────────────────────────────

% ── Game counts ───────────────────────────────────────────────────────────────
\newcommand{\NTotalGames}{23}
\newcommand{\NTotalGC}{19}          % ground_control games
\newcommand{\NTotalSF}{4}           % snowflake games
\newcommand{\NGamesClaude}{8}
\newcommand{\NGamesGemini}{6}
\newcommand{\NGamesGPT}{8}
\newcommand{\NGamesKimi}{1}

% ── Win rates ─────────────────────────────────────────────────────────────────
\newcommand{\WinsClaude}{2}
\newcommand{\WinsGemini}{1}
\newcommand{\WinsGPT}{0}
\newcommand{\WinsKimi}{0}
\newcommand{\WinRateClaude}{25.0}   % 2/8
\newcommand{\WinRateGemini}{16.7}   % 1/6
\newcommand{\WinRateGPT}{0.0}       % 0/8
\newcommand{\WinRateKimi}{0.0}      % 0/1

% ── Normalised score (agent / game-winner at final turn) ──────────────────────
% Computed across all admissible games per model
\newcommand{\NormScoreMedianClaude}{0.427}
\newcommand{\NormScoreMedianGemini}{0.481}
\newcommand{\NormScoreMedianGPT}{0.289}
\newcommand{\NormScoreMedianKimi}{0.478}    % n=1, no SD
\newcommand{\NormScoreMeanClaude}{0.400}
\newcommand{\NormScoreMeanGemini}{0.456}
\newcommand{\NormScoreMeanGPT}{0.346}
\newcommand{\NormScoreSDClaude}{0.147}
\newcommand{\NormScoreSDGemini}{0.165}
\newcommand{\NormScoreSDGPT}{0.225}

% ── Raw score at milestone turns (ground_control), mean ± SD ──────────────────
\newcommand{\TFiftyScoreClaude}{96.6}
\newcommand{\TFiftyScoreGemini}{112.2}
\newcommand{\TFiftyScoreGPT}{91.8}

\newcommand{\TFiftypmScoreClaude}{9.6}
\newcommand{\TFiftypmScoreGemini}{13.4}
\newcommand{\TFiftypmScoreGPT}{4.8}

\newcommand{\THundredFiftyScoreClaude}{394}
\newcommand{\THundredFiftyScoreGemini}{416}
\newcommand{\THundredFiftyScoreGPT}{320}

\newcommand{\THundredFiftypmScoreClaude}{86}
\newcommand{\THundredFiftypmScoreGemini}{83}
\newcommand{\THundredFiftypmScoreGPT}{84}

\newcommand{\TTwoHundredScoreClaude}{547}
\newcommand{\TTwoHundredScoreGemini}{572}
\newcommand{\TTwoHundredScoreGPT}{418}

\newcommand{\TTwoHundredpmScoreClaude}{110}
\newcommand{\TTwoHundredpmScoreGemini}{116}
\newcommand{\TTwoHundredpmScoreGPT}{112}

% ── Expansion — mean city count at milestone turns (ground_control) ───────────
\newcommand{\CityTFiftyClaude}{2.83}
\newcommand{\CityTHundredClaude}{4.67}
\newcommand{\CityTFiftyGemini}{2.67}
\newcommand{\CityTHundredGemini}{4.00}
\newcommand{\CityTFiftyGPT}{2.83}
\newcommand{\CityTHundredGPT}{4.67}
\newcommand{\CityTFiftyKimi}{2.00}
\newcommand{\CityTHundredKimi}{3.00}

% ── Tool calls per turn (mean across games, all turns) ────────────────────────
\newcommand{\MeanCallsPerTurnClaude}{10.7}
\newcommand{\MeanCallsPerTurnGemini}{20.8}
\newcommand{\MeanCallsPerTurnGPT}{18.7}
\newcommand{\MeanCallsPerTurnKimi}{15.6}
\newcommand{\SDCallsPerTurnClaude}{5.0}
\newcommand{\SDCallsPerTurnGemini}{9.0}
\newcommand{\SDCallsPerTurnGPT}{3.4}

% ── Aggregate tool call counts ───────────────────────────────────────────────
\newcommand{\NTotalToolCalls}{103703}

% ── ICC discriminative power (ground_control admissible) ─────────────────────
% Computed from analysis/figures/icc_table.tex (regenerated via plots.py)
\newcommand{\ICCExplorationTHundred}{0.717}     % discriminative
\newcommand{\ICCExplorationTHundredKWH}{12.28}  % Kruskal-Wallis H
\newcommand{\ICCExplorationTHundredKWP}{0.0065} % KW p-value
\newcommand{\ICCExplorationTFifty}{0.458}       % marginal

% ── Statistical tests ────────────────────────────────────────────────────────
\newcommand{\FisherChiSqPValue}{0.488}    % chi2 test on 4x2 win/loss table
\newcommand{\WinLossExactPValue}{0.530}
\newcommand{\KWScorePValue}{0.594}        % KW on normalised score
\newcommand{\KWScoreHStat}{1.90}          % KW on normalised score
\newcommand{\KWPMRHStat}{12.92}           % KW on between-model PMR
\newcommand{\KWPMRPValue}{0.002}          % significant
\newcommand{\SpearmanPMROutcome}{-0.012}  % PMR vs normalised score
\newcommand{\SpearmanPMROutcomeP}{0.957}
\newcommand{\BHFDRSurviving}{0}           % inflection delta tests surviving FDR
\newcommand{\BHFDRTotal}{267}             % total inflection delta tests

% ── Proactive Monitoring Rate — strategic_monitoring / denom calls per game ───
% PMR = mean(strategic_monitoring_calls / total_denom_calls) across games
\newcommand{\PMRClaude}{1.26}       % percent
\newcommand{\PMRGemini}{0.96}
\newcommand{\PMRGPTFive}{2.13}
\newcommand{\PMRKimi}{1.86}

% Victory-monitoring subcategory rate (get_victory_progress / denom)
\newcommand{\VictoryMonitoringRateClaude}{0.14}   % percent
\newcommand{\VictoryMonitoringRateGemini}{0.05}
\newcommand{\VictoryMonitoringRateGPT}{0.14}
\newcommand{\VictoryMonitoringRateKimi}{0.29}

% Mean get_victory_progress calls per game
\newcommand{\VictoryMonitoringCallsClaude}{4.2}
\newcommand{\VictoryMonitoringCallsGemini}{3.7}
\newcommand{\VictoryMonitoringCallsGPT}{7.4}
\newcommand{\VictoryMonitoringCallsKimi}{10.0}

% Total get_religion_spread calls across all admissible games
\newcommand{\NReligionSpreadCalls}{3}

% ── Missed-warning analysis (sensorium effect, §6.1) ─────────────────────────
% A "missed warning" = defeat with NO get_victory_progress call in the 20-turn
% window before game end. All defeats in this dataset have a detectable warning.
\newcommand{\NDefeats}{20}
\newcommand{\NDefeatsQueried}{12}        % at least one VP call in last 20 turns
\newcommand{\NDefeatsMissed}{8}          % no VP call in last 20 turns
\newcommand{\NDefeatsMissedClaude}{3}    % of 6 defeats
\newcommand{\NDefeatsMissedGemini}{1}    % of 5 defeats
\newcommand{\NDefeatsMissedGPT}{4}       % of 8 defeats
\newcommand{\NDefeatsMissedKimi}{0}      % of 1 defeat

% ── RAG validation (Cohen's κ between Haiku auto-labels and human labels) ────
\newcommand{\RAGKappa}{0.87}
\newcommand{\RAGKappaAgreement}{92.0}
\newcommand{\RAGKappaN}{100}

% ── Reflection-Action Gap (RAG@10) ────────────────────────────────────────────
% (Y + 0.5P) / total from analysis/rag_auto_results.csv (manifest: 13 admissible games, 35 turns)
\newcommand{\RAGClaude}{48.2}       % 20Y + 13P + 22N of 55
\newcommand{\RAGGemini}{65.8}       % 11Y + 3P + 5N of 19
\newcommand{\RAGGPTFive}{63.2}      % 38Y + 10P + 20N of 68
\newcommand{\RAGCommitmentsClaude}{55}
\newcommand{\RAGCommitmentsGemini}{19}
\newcommand{\RAGCommitmentsGPT}{68}
\newcommand{\RAGFollowedClaude}{20}
\newcommand{\RAGFollowedGemini}{11}
\newcommand{\RAGFollowedGPT}{38}

\newcommand{\RAGFiveClaude}{33.6}
\newcommand{\RAGTwentyClaude}{57.3}
\newcommand{\RAGFiveGemini}{50.0}
\newcommand{\RAGTwentyGemini}{68.4}
\newcommand{\RAGFiveGPT}{44.1}
\newcommand{\RAGTwentyGPT}{66.9}
\newcommand{\RAGPooledFive}{40.8}
\newcommand{\RAGPooledTen}{57.7}
\newcommand{\RAGPooledTwenty}{63.4}
\newcommand{\RAGTwentyNotExecuted}{27.5}
\newcommand{\RAGTwentyPartial}{18.3}

% RAG@10 95% CIs from per-commitment bootstrap (10,000 resamples, seed=0)
\newcommand{\RAGClaudeCILow}{36.4}
\newcommand{\RAGClaudeCIHigh}{60.0}
\newcommand{\RAGGeminiCILow}{47.4}
\newcommand{\RAGGeminiCIHigh}{84.2}
\newcommand{\RAGGPTCILow}{52.2}
\newcommand{\RAGGPTCIHigh}{73.5}

% ── ELO leaderboard (Appendix E) ──────────────────────────────────────────────
% FFA pairwise scoring, K = 32/sqrt(N-1), base 1500. Computed by analysis/elo.py
% over the 23 admissible games (winnerCiv normalised so the Kimi-K2.5 game
% with "CIVILIZATION_JAPAN" is included; the production dashboard drops it).
\newcommand{\EloClaude}{1547}
\newcommand{\EloGemini}{1511}
\newcommand{\EloGPTFive}{1457}
\newcommand{\EloKimi}{1496}
% Top AI leaders by ELO
\newcommand{\EloHojoTokimune}{1662}
\newcommand{\EloTomyris}{1583}
\newcommand{\EloSeondeok}{1570}
\newcommand{\EloRobertTheBruce}{1569}

\maketitle

\begin{abstract}
We present CivBench, an open-source benchmark for evaluating language model agents in long-horizon, tool-mediated environments through the Model Context Protocol (MCP). A single episode can span over 300 turns and produces thousands of tool calls over a large action space, requiring sustained planning, state monitoring, and execution under partial observability. The environment exposes 76 MCP tools and a deterministic narration layer that retrieves and formats game state as structured text.

We use CivBench to characterise agent behaviour across four model families in \NTotalGames{} admissible runs. The sample is a pilot, not a model ranking: aggregate outcomes do not reliably discriminate models at this scale. Instead, we introduce two interface-level metrics that the environment makes measurable: Proactive Monitoring Rate (PMR), capturing whether agents actively query latent strategic state, and RAG@10, capturing whether commitments stated in structured planning reflections are executed within ten subsequent turns.

Across runs we observe two consistent patterns under a shared playbook protocol. Agents under-monitor strategically relevant state that is available but requires explicit querying: despite playbook guidance to query victory progress every 20 turns, agents do so only every 30–75 turns, and in \NDefeatsMissed{} of \NDefeats{} detectable defeats they failed to query within the 20-turn warning window before game end. Agents also frequently fail to execute near-term commitments stated in their own planning reflections (RAG@10 between \RAGClaude{}\% and \RAGGemini{}\% across models). Both patterns arise despite tool access and explicit guidance, and we interpret them as deviations under instruction rather than absences of capability.

We release the environment, scenarios, logs, metrics, and analysis pipeline at \url{https://github.com/lmwilki/civ6-mcp}.
\end{abstract}

\section{Introduction}
\label{sec:intro}

Frontier models are increasingly deployed as agents: they call tools, inspect external state, and act over extended sequences of decisions. Evaluating this behaviour requires environments that capture sustained operation rather than isolated question answering, including partial observability, large action spaces, and decisions whose effects may only appear much later.

Existing evaluations often focus on individual components of this process, such as reasoning, tool use, or short-horizon planning, but provide limited visibility into how these capabilities interact over long horizons. In particular, evaluating whether agents maintain awareness of relevant information and translate stated plans into subsequent actions requires access to their behaviour throughout extended interactions.

Civilization~VI (Fig.~\ref{fig:game_screenshot}) provides a suitable testbed for this setting. Games can span over 300 turns and require simultaneous management of economic, scientific, cultural, military, diplomatic, spatial, and temporal priorities. The game offers six
victory conditions, with decisions across different systems interacting and often having consequences many turns later. While environments such as Diplomacy, Minecraft, and Craftax capture important aspects of long-horizon agency, Civilization~VI brings these demands together
in a persistent world where early decisions constrain future options. Agents must track plans, monitor rivals, manage competing objectives, and revise strategy as new information arrives. This makes Civilization~VI useful for evaluating situated, tool-mediated behaviour rather than static game knowledge.

We present CivBench, an open-source benchmark that connects language model agents to Civilization~VI through the Model Context Protocol (MCP). CivBench exposes 76 MCP tools covering core game systems, including state queries, unit control, city management, diplomacy, and research. It also provides a deterministic narration layer that retrieves and formats game state as structured text, without an additional language model. Agents interact with the game through MCP tool calls rather than pixels.

The central design choice in CivBench is to separate availability of information from whether it is brought into context. Non-local state must be accessed through explicit queries, allowing us to distinguish between information that is unavailable and information that is available but not retrieved. This makes it possible to examine how agents allocate attention across different information sources and whether their stated plans are followed by corresponding actions.

Beyond evaluating game outcomes, CivBench provides a setting for investigating how agent design affects long-horizon behaviour. Researchers can compare models and scaffolding approaches, including memory mechanisms, planning strategies, and monitoring protocols, under controlled scenarios. Complete interaction traces allow these behaviours to be studied directly, including whether agents retrieve important information, maintain strategic objectives, and execute their stated commitments. This supports the investigation of failure modes and potential improvements that may not be apparent from aggregate performance alone.

We evaluate four model families across \NTotalGames{} admissible full-game runs. With this sample size, the data does not support stable model ranking. Instead, we analyse behaviour in the traces using two interface-level metrics: Proactive Monitoring Rate (PMR), measuring the proportion of tool calls devoted to strategic monitoring, and Reflection--Action Gap (RAG@K), measuring the execution of stated commitments within a subsequent window (formally defined in Section~4.2). These metrics expose aspects of agent behaviour that are not captured by game outcomes alone.

\paragraph{Key contributions:}

\begin{figure}[t]
\centering
\includegraphics[width=\textwidth]{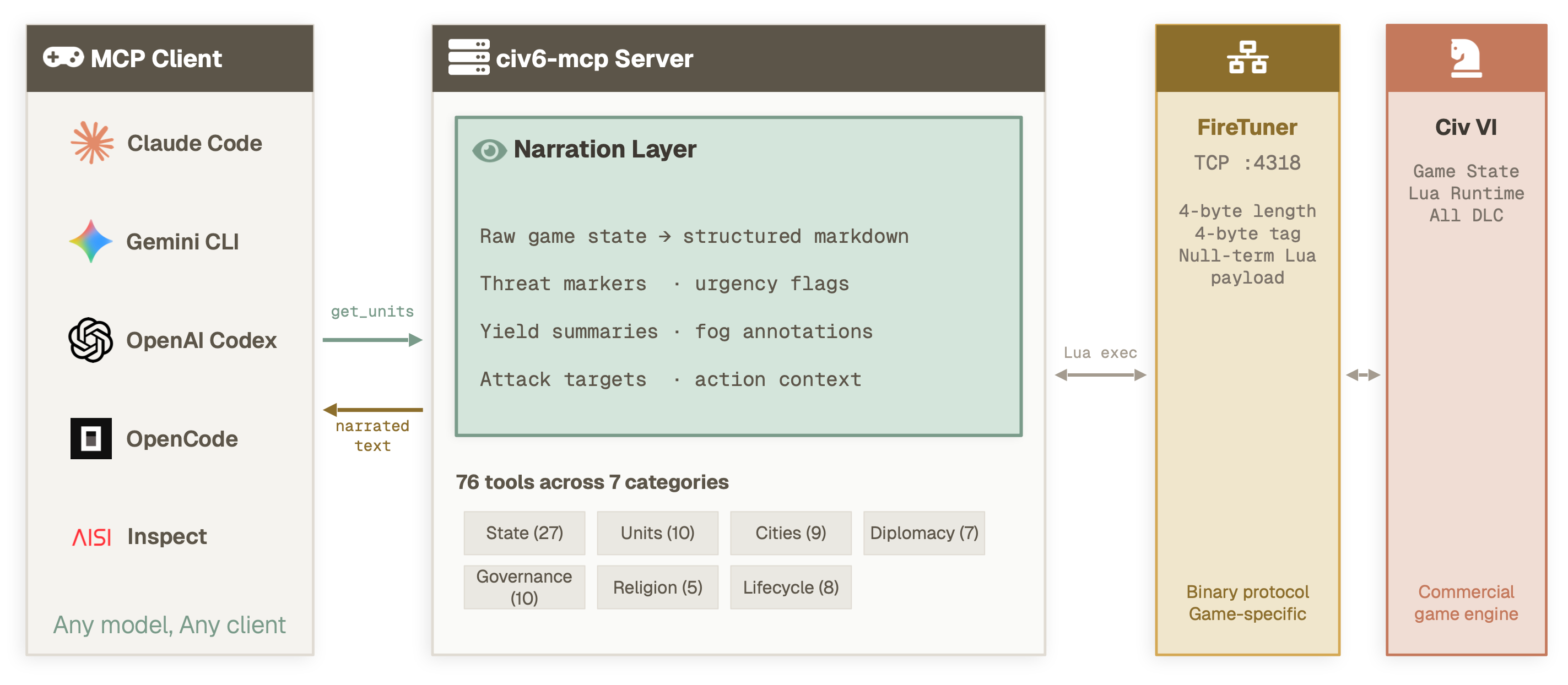}
\caption{Model Context Protocol (MCP) architecture. Agents interact via MCP tools; the server translates calls to Civilization~VI and returns structured observations through the narration layer. (Sec.~\ref{sec:environment})}
\label{fig:architecture}
\end{figure}

\begin{enumerate}

\item \textbf{CivBench benchmark environment.} We introduce an MCP benchmark for full-game Civilization~VI play with 76 tools, fixed scenarios, and structured logs (Fig.~\ref{fig:architecture}).

\item \textbf{Narration protocol.} We design a deterministic interface that converts game state into structured, agent-readable observations while preserving the requirement that relevant information must be explicitly queried, enabling controlled observability.

\item \textbf{Behavioural metrics.} We introduce Proactive Monitoring Rate (PMR), which measures the allocation of tool calls to strategic monitoring, and RAG@K, which measures whether stated commitments are executed within a subsequent window.

\item \textbf{Empirical characterisation.} Across \NTotalGames{} runs spanning four model families, we demonstrate that the environment and metrics make two interface-level patterns measurable: under-monitoring of queryable global state and failure to execute near-term commitments. We frame these as pilot findings to be reproduced at scale, not as model rankings.

\end{enumerate}

\section{Related Work}

\label{sec:related}

\subsection{Civilization as an AI Environment}

Prior Civilization-based environments differ in engine and interface. CivRealm~\citep{qi2024} uses FreeCiv with a Gymnasium-style API, while CivAgent~\citep{fuxiailab2024} and Vox Deorum~\citep{voxdeorum2025} study LLM behaviour in alternative implementations of the game.

CivBench differs in three ways (Tab.~\ref{tab:civrealm}). First, it uses MCP rather than Gymnasium, matching production tool interfaces. Second, it introduces a narration layer that separates availability of state from whether it is explicitly queried, enabling controlled observability. Third, it targets Civilization~VI, whose richer, more well-documented mechanics allow us to examine the gap between what an agent \textit{knows} and what it \textit{does} (Section~\ref{sec:sensorium}).

\subsection{LLM Evaluation}

Recent work evaluates LLM agents in interactive environments with varying scope. BALROG~\citep{paglieri2025balrogbenchmarkingagenticllm} reports a gap between explaining strategies and executing them but does not use MCP. CICERO~\citep{fair2022} demonstrates strong performance in Diplomacy, while shorter-horizon game benchmarks~\citep{costarelli2024, duan2024} and multi-agent frameworks~\citep{rutherford2024,ellis2023} study related capabilities under more constrained settings.

Other benchmarks isolate specific components of agent behaviour, including long-term management (Vending-Bench~2~\citep{andon2025}), tool use (MCPAgentBench~\citep{mcpagentbench2025}, ToolBench, API-Bank~\citep{qin2023toolllmfacilitatinglargelanguage,li2023apibankcomprehensivebenchmarktoolaugmented}), and software engineering (SWE-bench~\citep{jimenez2024}). 

CivBench integrates these elements in a single long-horizon, tool-mediated environment with a production-style MCP interface, enabling analysis of attention allocation and plan execution through interaction traces rather than aggregate outcomes. Unlike prior game-agent benchmarks that primarily score task success or final outcomes, CivBench releases full tool-call traces and defines metrics over information retrieval and commitment execution.
\section{Environment}
\label{sec:environment}

\subsection{Civilization VI}

Civilization~VI is a 4X strategy game in which players explore, expand, exploit resources, and compete with rival civilisations. Games can span over 300 turns and require decisions across economic, scientific, cultural, military, diplomatic, spatial, and temporal domains (List~\ref{list:domains}). The game offers six victory conditions---Science, Culture, Domination, Religion, Diplomacy, and Score---requiring agents to balance competing priorities under resource constraints. For example, a Science victory requires sustained technological development followed by late-game space projects, involving research, production, and economic planning.

Civilization~VI is also useful for separating game knowledge from execution. While strategy knowledge may appear in training data, each run presents a new state with imperfect information and many tool-mediated decisions. CivBench therefore evaluates behaviour through interaction with a live game state rather than memorised strategy alone.

\subsection{The MCP Interface}

CivBench connects to Civilization~VI through the FireTuner protocol over TCP. The CivBench server exposes 76 MCP tools covering core systems, including state queries, unit actions, city management, diplomacy, research, governance, religion, trade, and game lifecycle operations (Tab.~\ref{tab:tools}).

Each turn involves multiple tool calls, resulting in thousands of interactions over a full game. The agent must decide which tools to call, when to call them, and how to act on the returned text. This interface resembles real-world tool use more than a Gymnasium-style API, where interaction is typically mediated through fixed observation spaces.

\subsection{The Narration Layer}

The narration layer is the main interface contribution of CivBench. The agent only observes state it explicitly queries: relevant information is representable but not passively observed and must be actively retrieved through tool calls.

The narration layer is implemented entirely through deterministic code, without an additional language model or learned component. State-query tools retrieve game state through Lua queries, which are parsed and formatted into structured text using Python. Some outputs also include fixed, rule-based summaries, such as assessments of victory path viability. The same query in the same game state therefore returns the same observation, although subsequent game trajectories need not be deterministic.

This design separates availability of information from whether it is brought into context. As a result, CivBench can distinguish between information that is unavailable and information that is available but not queried, enabling analysis of attention allocation through tool-call traces.

Observations are returned as structured text rather than raw tensors or pixels, with formatting and annotations that support decision-making while preserving the requirement that the agent must decide what to query. The narration layer comprises 37 functions spanning core game systems (Tab.~\ref{tab:narration}).

Crucially, the interface improves representation but does not enforce retrieval: relevant state (e.g., victory progress) produces no observation unless explicitly queried. This boundary underlies the sensorium analysis in Section~\ref{sec:sensorium}. The choice of representation may nevertheless influence agent behaviour, and its effects have not been isolated through a controlled ablation.

CivBench is intended primarily as a benchmark artifact rather than a leaderboard in this release. A benchmark instance consists of a fixed save/scenario, model configuration, shared playbook, complete MCP transcript, and derived metric table. The released scripts recompute outcome metrics, tool-use profiles, PMR, and RAG@K from raw logs. This makes the current 23-game dataset a pilot reference set, while the main contribution is the reusable environment and trace-level measurement protocol.
\section{Evaluation Framework}
\label{sec:evaluation}

\subsection{Experimental Design}

We evaluate a single LLM agent against Civilization~VI's built-in AI opponents. Model identifiers, endpoints, access dates, and harness versions are reported in Appendix~\ref{sec:models_access} and Table~\ref{tab:model-summary}. Each scenario is defined by fixed map and game seeds. Reproducibility also depends on the game version, DLC configuration, scenario files, model configuration, and playbook version.

\begin{table}[t]
\centering
\caption{\textbf{Evaluation scenarios} increasing in difficulty.}
\label{tab:scenarios}
\small
\begin{tabular}{lllll}
\toprule
\textbf{Scenario} & \textbf{Map Type} & \textbf{Speed} & \textbf{Difficulty} & \textbf{Purpose} \\
\midrule
Ground Control & Pangaea, standard & Quick & Prince & Baseline competence \\
Snowflake & Six-arm snowflake, small & Quick & King & Military focus \\
Cry Havoc & Pangaea, tiny & Quick & Immortal & Stress test \\
\bottomrule
\end{tabular}
\end{table}

\paragraph{Scenario rationale.}

Ground Control is the baseline scenario. Snowflake restricts play to military victory on an isolated map. Cry Havoc introduces a severe disadvantage and is not included in the reported results. Because Snowflake has only four admissible runs and Cry Havoc is excluded, scenario-level conclusions are descriptive.

\paragraph{Agent playbook.}

All models receive the same versioned playbook, which specifies turn structure, checkpoints, and a five-field diary (tactical, strategic, tooling, planning, hypothesis). Earlier versions used hard triggers; the reported runs use a softer advisory version. The playbook standardises interaction structure while leaving tool selection and prioritisation to the agent.

\subsection{Metrics}
\label{sec:metrics}

We report both outcome measures and behavioural metrics. Normalised score captures competitive standing, while PMR and RAG@K measure monitoring and execution behaviour not visible in aggregate outcomes.

\paragraph{Normalised score.}

\texttt{normalised\_score = agent\_raw\_score / winner\_raw\_score\_at\_game\_end}. Winning agents score 1.0; others reflect relative standing. We treat this as descriptive rather than a measure of strategic competence, since score aggregates multiple systems into a single outcome.

\paragraph{PMR: Proactive Monitoring Rate.}

\texttt{PMR = strategic\_monitoring\_tool\_calls / all\_non\_infrastructure\_tool\_calls}. Monitoring tools require proactive queries (e.g. \texttt{get\_victory\_progress}, \texttt{get\_diplomacy}), while infrastructure calls (e.g. \texttt{end\_turn}) are excluded. PMR measures how much of the tool budget is allocated to maintaining a global view rather than reacting to local state, providing a proxy for attention allocation. A higher PMR is not necessarily better, as additional monitoring may be unnecessary or inefficient. The relevant finding is that agents under-monitor explicitly recommended signals, sometimes failing to retrieve detectable threats. We do not establish that increasing PMR would improve game outcomes.

\paragraph{RAG@K: Reflection--Action Gap.}

\texttt{RAG@K = (Y + 0.5P) / total\_commitments}, where commitments are extracted from planning reflections and evaluated over $K$ subsequent turns ($K=10$). A commitment specifies a target (e.g. unit, city, or research item) and is labelled as "executed" (Y), "partial" (P), or "not executed". Commitments are extracted and labelled using a fixed LLM-assisted rubric, with human validation described in Section~\ref{sec:ragap}. RAG@K measures the extent to which stated plans are translated into subsequent actions, with higher values indicating greater execution fidelity.

We use a fixed horizon ($K=10$) because the playbook asks agents to plan concrete actions over the next 5--10 turns, while a relatively short evaluation window limits ambiguity from changing circumstances. Sensitivity to the evaluation horizon is examined in Section~\ref{sec:ragap}. RAG@K measures whether commitments are executed as stated, but cannot distinguish poor follow-through from appropriate replanning when circumstances change. It should therefore be interpreted as execution fidelity for concrete, self-stated near-term commitments under a shared diary protocol, not as a complete measure of planning quality.

Both PMR and RAG@K are interface-level metrics capturing monitoring and execution behaviour under tool-mediated interaction.

\subsection{Practical Considerations}

Civilization~VI is commercially licensed; users must own a copy to conduct new game runs. Reproducibility depends on the pinned game version, configuration, seeds, and playbook (Appendix~\ref{sec:reproducibility}). CivBench integrates with the Inspect framework \citep{aisi2024} for execution, logging, and metric extraction. The released traces and analysis pipeline allow researchers to reproduce the reported metrics and figures without owning or running Civilization~VI.

\subsection{Scope, Limitations, and Threats to Validity}
\label{sec:limitations}

The dataset provides evidence for two behavioural patterns: agents under-monitor latent state and frequently fail to execute stated plans. It also demonstrates how aggregate outcome measures can obscure variation in agent behaviour. These observations are grounded in interaction traces under a fixed protocol and should be interpreted within that scope.

\paragraph{Sample size and power.} With \NTotalGames{} runs across four model families, the sample is small and uneven, limiting statistical power. Between-model differences are therefore descriptive rather than reliable estimates of relative performance.

\paragraph{Playbook confound.} The shared playbook introduces human guidance, so behaviour reflects the combination of model reasoning, tool use, and instruction following. Failures to act on explicitly recommended signals should be read as deviations under guidance rather than absence of capability. The playbook was introduced in part to improve evaluation reliability: only 4 of 19 pre-harness development runs (21\%) reached a natural conclusion (Fig.~\ref{fig:harness_comparison}). These earlier runs do not constitute a controlled playbook-free comparison.

\paragraph{Baselines.} We do not include a random, scripted, or heuristic baseline operating through the same MCP interface. Such a baseline would help calibrate monitoring behaviour, tool allocation, and failure rates without necessarily requiring competitive gameplay. The built-in Civilization~VI opponents and pre-harness development runs do not provide this comparison because they do not operate under the same interface and evaluation protocol.

\paragraph{Environmental constraints.} Civilization~VI introduces version and licensing dependencies; the FireTuner protocol supports a single connection (limiting multi-agent setups). Each full-game run costs approximately \$31--229 in API fees and takes 2--8 hours on a local consumer machine running Civilization~VI. Results cover Ground Control and Snowflake. All observed victories occur on Ground Control; behaviours may vary under different maps, difficulties, or victory constraints.

\paragraph{Contamination.} Strategy knowledge may appear in training data, but each run requires live adaptation under partial observability, shifting the evaluation towards interaction behaviour rather than memorised strategy.

\paragraph{Measurement scope.} PMR and RAG@K depend on the defined set of monitoring tools and the interpretation of planning text. PMR does not establish whether additional monitoring would improve outcomes, while RAG@K may penalise appropriate changes of plan and depends on LLM-assisted commitment labelling. Results should be read as measurements under the CivBench protocol rather than exhaustive characterisations of agent capability.
\section{Results: Descriptive Behavioural Profiles}
\label{sec:results}

Our dataset comprises \NTotalGames{} admissible games across Ground Control (\NTotalGC{}) and Snowflake (\NTotalSF{}), drawn from four model families with uneven coverage: three families have 6–8 runs and one exploratory family has a single run. The sample is small and uneven; we therefore present results as descriptive characterisations of behaviour rather than model comparisons, and focus on patterns that are consistent across runs.

\subsection{Aggregate Outcomes Are Insufficient at This Scale}
\label{sec:results-aggregate}

Aggregate outcome measures do not discriminate between models in this dataset. Win/loss counts are sparse, with only 3 victories in 23 admissible runs, all on Ground Control; Fisher--Freeman--Halton exact test does not support between-model discrimination (p = 0.530; Fig. ~\ref{fig:outcome_heatmap}). Normalised score is also weakly discriminating ($H=\KWScoreHStat{}$, $p=\KWScorePValue{}$; Fig.~\ref{fig:normalised_score}), with within-model variance comparable to between-model differences. All observed wins were science victories on Ground Control; no model won on Snowflake. Apparent differences in raw counts are therefore not interpreted as evidence of model superiority.

This pattern extends across most aggregate measures. An exploratory ICC analysis on Ground Control runs (Tab.~\ref{tab:icc_table}) finds that of 13 candidate metrics, exploration at T100 is the strongest exploratory signal (ICC=0.717); final score, city counts, and economic yields all have ICC near zero or negative, indicating that within-model variance dominates. This is consistent with the sample size but also reflects a substantive point: outcome-level measures aggregate over hundreds of decisions, compressing the behavioural variation that distinguishes these agents. The remainder of the paper therefore turns to interface-level metrics that expose this variation directly.

\subsection{Tool-Use Profiles}

Across the pilot runs, tool budgets vary substantially, but local actions and state queries dominate, while strategic monitoring remains a small fraction of activity across models. CivBench records how agents allocate tool calls over long horizons, making this behaviour observable through interaction traces. Section~\ref{sec:sensorium} examines the associated information-retrieval failures.

\begin{figure}[t]
\centering
\includegraphics[width=\textwidth]{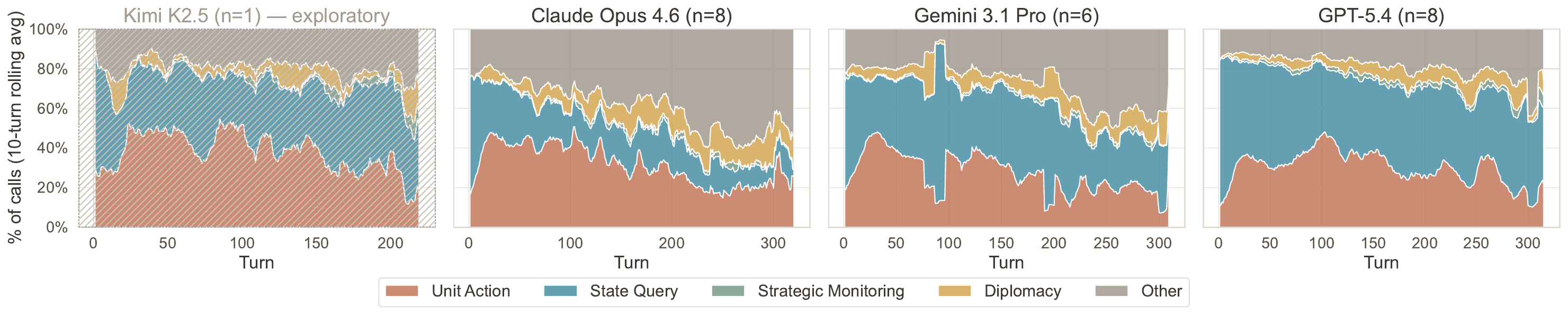}
\caption{\textbf{Tool call composition over 10-turn rolling average.} Tool budgets are dominated by local actions (red) and state queries (blue); strategic monitoring remains consistently low across models.}
\label{fig:tool_stacked_area}
\end{figure}
\section{Analysis}
\label{sec:analysis}

\subsection{The Sensorium Effect}
\label{sec:sensorium}

The clearest pattern in the traces is the \textit{sensorium effect}: the gap between passive human perception and active agent queries. A human player absorbs many signals without effort, while an agent only observes what it explicitly queries. Each query consumes time and context. This is therefore an attention-allocation problem rather than a tooling gap.

\begin{figure}[t]
\centering
\includegraphics[width=\textwidth]{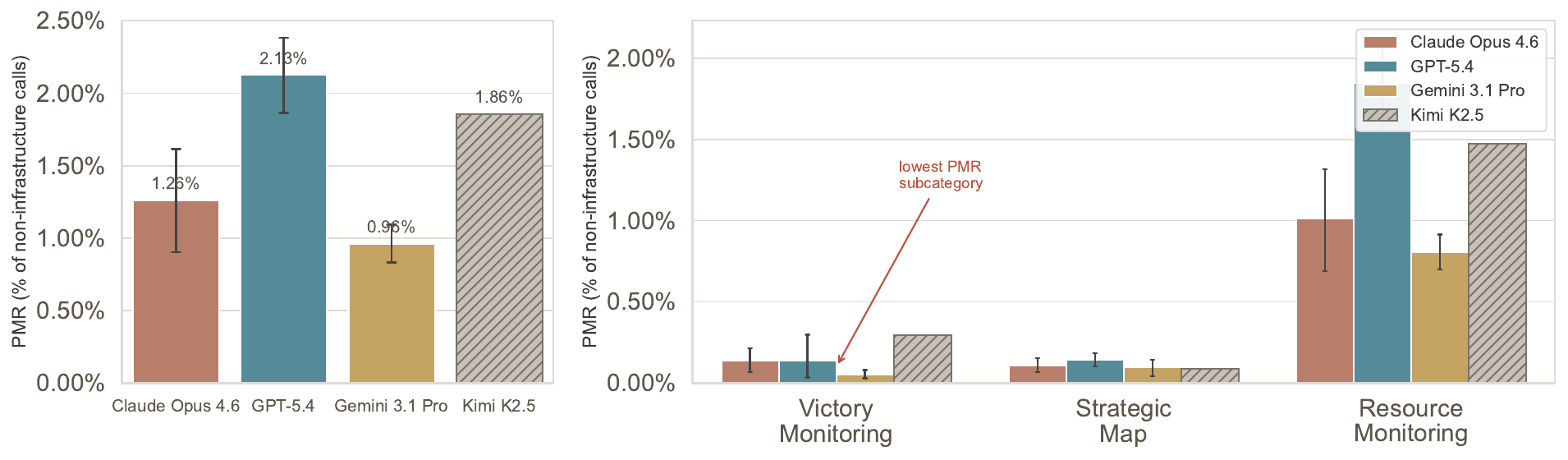}
\caption{\textbf{Descriptive Proactive Monitoring Rate by model (A) and subcategory (B).} Strategic monitoring tool calls divided by all non-infrastructure tool calls. Monitoring rates are uniformly low across models, with particularly sparse querying of victory progress, indicating that globally relevant state is rarely brought into context despite availability and guidance.}
\label{fig:pmr_subcategories}
\end{figure}

\begin{table}[ht]
\centering
\caption{Missed-warning analysis for defeated runs. A warning is counted as detectable if \texttt{get\_victory\_progress} would have exposed a rival victory threat at least 20 turns before game end. Queried = at least one \texttt{get\_victory\_progress} call in the 20-turn window before game end. Missed = detectable defeats with no query in the window.}
\label{tab:missed-warnings}
\begin{tabular}{lrrrr}
\toprule
Model & Defeats & Detectable & Queried & Missed \\
\midrule
Claude Opus 4.6 & 6 & 6 & 3 & \NDefeatsMissedClaude{} \\
Gemini 3.1 Pro & 5 & 5 & 4 & \NDefeatsMissedGemini{} \\
GPT-5.4 & 8 & 8 & 4 & \NDefeatsMissedGPT{} \\
Kimi-K2.5 & 1 & 1 & 1 & \NDefeatsMissedKimi{} \\
\midrule
\textbf{Total} & \textbf{\NDefeats{}} & \textbf{\NDefeats{}} & \textbf{\NDefeatsQueried{}} & \textbf{\NDefeatsMissed{}} \\
\bottomrule
\end{tabular}
\end{table}

We quantify this behaviour using Proactive Monitoring Rate (PMR) (Section~\ref{sec:metrics}, Fig.~\ref{fig:pmr_subcategories}). Aggregate PMR is low in every tested model family, ranging from 0.96\% to 2.13\% of non-infrastructure calls. Victory monitoring is particularly sparse, at \VictoryMonitoringRateGemini{}--\VictoryMonitoringRateKimi{}\% of calls, corresponding to \VictoryMonitoringCallsGemini{}--\VictoryMonitoringCallsKimi{} \texttt{get\_victory\_progress} calls per game. Despite playbook guidance to check every 20 turns, agents query this signal only once every 30--75 turns.

Given the small and uneven sample, we treat differences between models as descriptive and focus on the consistent pattern across runs: monitoring remains a small fraction of the tool budget.

The consequence is that agents can miss important information even when it is available. Table~\ref{tab:missed-warnings} shows that in \NDefeatsMissed{} of \NDefeats{} detectable defeats, agents failed to query victory progress within the 20-turn warning window. In these cases, the relevant information was available through the interface but was not retrieved, regardless of how the narration layer would have presented it. This establishes a failure to retrieve available strategic information, although it does not establish that additional monitoring would have prevented the defeats.

PMR does not increase towards the endgame (Fig.~\ref{fig:pmr_over_time}), despite the potential importance of monitoring as victory or defeat approaches. This suggests that sparse strategic monitoring persists throughout the game rather than being confined to earlier stages.

More broadly, this pattern arises from the interface constraint that relevant state must be actively queried. As a result, performance depends not only on reasoning over available information, but on whether that information is retrieved at all. While demonstrated here in Civilization~VI, similar dynamics may arise in other tool-mediated settings where critical state is not passively observed.

\subsection{The Reflection--Action Gap}
\label{sec:ragap}

We define the Reflection--Action Gap (RAG) as the tendency of agents to state concrete commitments in planning reflections without executing them. RAG@10 (Section~\ref{sec:metrics}) scores execution fidelity over the subsequent 10 turns, assigning full credit to executed commitments and half credit to partially executed commitments; higher scores indicate better follow-through.

Examples from the traces include: \textit{"Build campuses in new cities"}, \textit{"Found second city"}, and \textit{"Check victory progress"}. These commitments were not executed within the next 10 turns.

Commitments were extracted and labelled by Claude Haiku 4.5 using a fixed rubric. We validated the pipeline on a stratified random sample of \RAGKappaN{} commitments drawn proportionally across the three models with sufficient runs (Claude, Gemini, GPT), labelled independently by an author. Agreement with the LLM labels was \RAGKappaAgreement{}\% (Cohen's $\kappa=\RAGKappa{}$). The labeller (Haiku 4.5) shares a model family with one evaluated agent (Claude Opus 4.6), introducing a potential source of same-family evaluation bias \citep{spiliopoulou2025playfavoritesstatisticalmethod}. Claude has the lowest RAG@10 score of the three evaluated families, which does not suggest favourable same-family bias in these aggregate results but does not rule it out.

\begin{figure}[t]
\centering
\includegraphics[width=\textwidth]{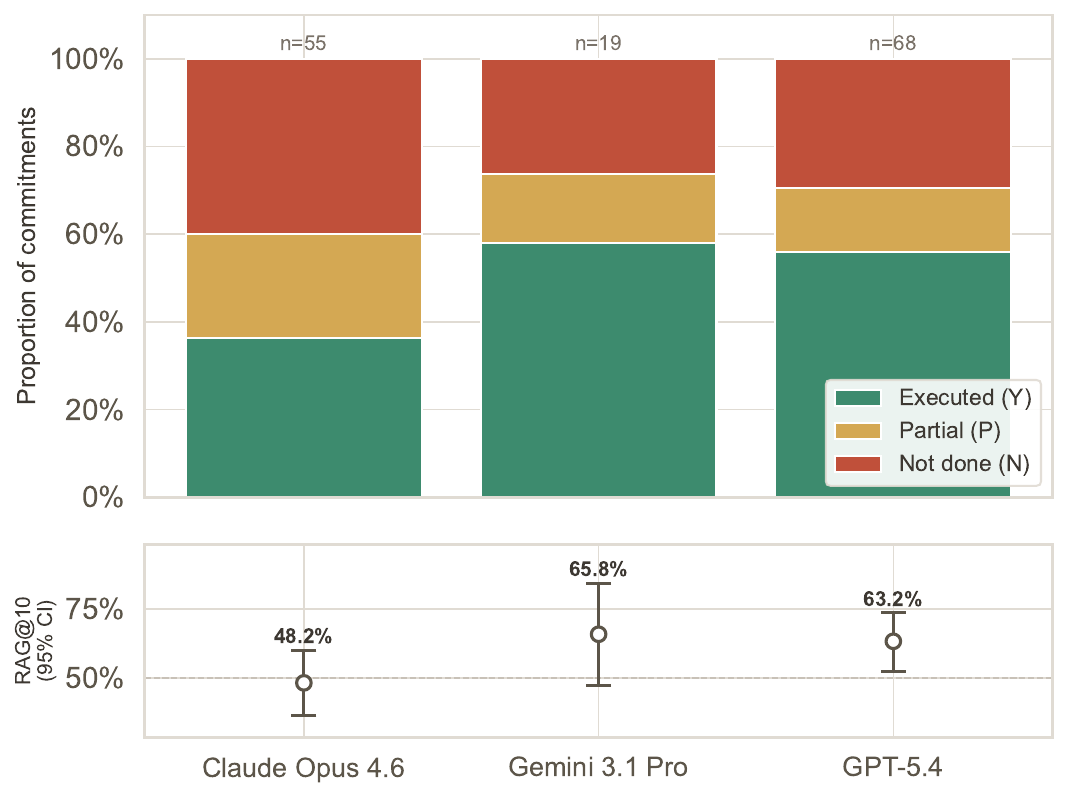}
\caption{\textbf{Commitment execution within 10 turns (RAG@10).} Fraction of commitments executed (Y), partially executed (P), or not executed (N). Error bars show 95\% confidence intervals from 2,000 commitment-level bootstrap resamples, without accounting for within-game dependence. A substantial fraction of commitments are not executed within the evaluation window across tested models, indicating a consistent gap between stated plans and subsequent actions.}
\label{fig:rag_breakdown}
\end{figure}

RAG@10 ranges from \RAGClaude{}\% to \RAGGemini{}\% across the three model families with sufficient commitments for analysis, with overlapping bootstrap confidence intervals (Fig.~\ref{fig:rag_breakdown}). A substantial fraction of commitments are not executed within the next ten turns. This pattern is consistent with prior findings that models can explain strategies without executing them \citep{paglieri2025balrogbenchmarkingagenticllm,schmied2025}; CivBench makes the behaviour directly observable through diary entries and tool-call traces. The analysis covers 142 commitments from 27 selected diary turns across 13 runs, with sampling details provided in Appendix~\ref{sec:rag_sampling}.

\paragraph{Horizon sensitivity.} We evaluate RAG at $K=5$, $K=10$, and $K=20$, obtaining pooled execution-fidelity scores of \RAGPooledFive{}\%, \RAGPooledTen{}\%, and \RAGPooledTwenty{}\%, respectively. The execution gap persists across all three horizons (Appendix~\ref{sec:rag_sensitivity}).

\paragraph{Limitations.} RAG depends on an LLM-assisted pipeline that extracts commitments from diary text, applied uniformly across the analysed runs and assessed against a human-labelled validation subset ($\kappa=\RAGKappa{}$). Although human validation shows substantial agreement, evaluator bias remains possible. Moreover, RAG measures whether commitments are executed as stated, not whether executing them remains appropriate as circumstances change. It therefore cannot distinguish poor follow-through from sensible replanning and may penalise commitments that become infeasible. Results should be interpreted as measurements of near-term execution fidelity under the shared diary and labelling protocol, rather than ground truth about planning quality.
\section{Discussion}
\label{sec:discussion}

\subsection{Research Benefits}

CivBench supports two complementary approaches to studying long-horizon agent behaviour. First, researchers can evaluate models and scaffolding approaches, including memory mechanisms, planning strategies, and monitoring protocols, under fixed scenarios and compare both game outcomes and trace-level behaviour. Second, the released trajectories and analysis pipeline allow researchers to reproduce the reported metrics, develop alternative measures, and investigate failure modes without rerunning Civilization~VI. By separating information availability from retrieval, CivBench provides a setting for investigating how agents monitor their environment and execute stated commitments. The current evaluation is a pilot demonstration of these capabilities rather than a comprehensive comparison of agent designs.

\subsection{Structured Reflection}

We implement a turn diary as both an intervention and a measurement mechanism. The diary requires five fields at each turn (\texttt{tactical}, \texttt{strategic}, \texttt{tooling}, \texttt{planning}, \texttt{hypothesis}) and can be retrieved through \texttt{get\_diary}. This is intended to mitigate context loss in long episodes.

The diary also enables RAG by providing explicit commitments comparable to subsequent actions. This introduces a measurement confound: the playbook elicits explicit planning, so RAG measures execution of prompted commitments rather than spontaneous plans. We accept this trade-off because explicit commitments are more interpretable than intent inferred from actions alone. Earlier pre-harness runs were not controlled ablations of the playbook or diary (Section~\ref{sec:limitations}). RAG should therefore be read as execution fidelity under a shared elicitation protocol, not as an estimate of unprompted planning ability.

Whether structured reflection improves or degrades execution remains open. Future work should test this through ablations, including removing the diary, introducing persistent commitment tracking, and enforcing monitoring schedules.

\subsection{Design Implications}

The observed failure modes suggest that improving reasoning alone may be insufficient for reliable long-horizon behaviour. Monitoring could be supported by scheduled queries or interfaces that proactively surface critical information, while persistent memory, task queues, and explicit commitment tracking may improve execution fidelity. These remain hypotheses requiring controlled evaluation. By making information retrieval and commitment execution measurable, CivBench provides a setting in which such interventions can be tested alongside conventional outcome metrics.
\section{Conclusion}
\label{sec:conclusion}

We introduced CivBench, an open-source benchmark for evaluating long-horizon, tool-mediated agents in Civilization~VI through the Model Context Protocol. By separating information availability from retrieval and recording complete interaction traces, CivBench enables analysis of monitoring behaviour and the execution of stated plans beyond aggregate game outcomes.

Across 23 pilot runs, we identify two patterns under a shared playbook: agents frequently fail to retrieve strategically relevant information and leave concrete near-term commitments unexecuted. These findings do not establish general capability deficits or reliable model rankings, but demonstrate the value of trace-level evaluation. We release the environment, scenarios, logs, metrics, and analysis pipeline to support further research into reliable long-horizon agency.
\section{Acknowledgements and Disclosures}
\label{sec:acknowledgements}

ATDA is supported by the EPSRC Centre for Doctoral Training in Health Data Science (EP/S02428X/1). This work was supported by the EPSRC (grant UKRI5149) and other sponsors of BOLD (British Open-ended Learning and Discovery Lab). None of the authors has competing interests to declare.

\FloatBarrier
\newpage
\bibliographystyle{plainnat}
\bibliography{references}

\newpage

\appendix
\FloatBarrier

\section{Technical Details}

\label{app:technical}

A run is admissible if it uses the pinned scenario/playbook/harness version, reaches a natural game conclusion or the scenario turn limit, and has complete tool-call logs sufficient to recompute all metrics.

% Placeholder for full technical specification.

\subsection{Domains}

\label{sec:domains}

\begin{enumerate}

\label{list:domains}

\item \textbf{Economic.} Gold income, trade routes, city yields, luxury and strategic resource stockpiles.

\item \textbf{Scientific.} A technology tree of 67 technologies across 8 eras, with eureka boosts rewarding in-game achievements.

\item \textbf{Cultural.} A civic tree, government selection, policy card optimisation, and Great Works.

\item \textbf{Military.} Unit production and composition, hex-grid positioning, combat strength calculations, promotions, and upgrade paths.

\item \textbf{Diplomatic.} Bilateral relationships, grievance tracking, delegations, embassies, trade deals, World Congress voting, and formal alliances.

\item \textbf{Spatial.} A hex grid of approximately 4,000 tiles with procedurally generated terrain, features, rivers, resources, fog of war, district adjacency bonuses, and city spacing constraints.

\item \textbf{Temporal.} Multi-turn production queues, research timelines, growth projections, era progression, and irreversible commitment decisions (district placement, government selection, city founding).

\end{enumerate}

\begin{table}[htbp]

\centering

\caption{Narration functions by domain. The full narration layer comprises 37 functions.}

\label{tab:narration}

\small

\begin{tabular}{clp{7.5cm}}

\toprule

\textbf{Domain} & \textbf{Fn} & \textbf{What the agent sees} \\

\midrule

Overview      & 1 & Turn, yields, score, era, rankings, exploration \% \\

Units         & 4 & Position, HP, moves, charges, attack targets, upgrades, threats \\

Cities        & 3 & Population, yields, production queue, walls, growth, loyalty, districts \\

Map / Spatial & 3 & Hex tiles, strategic fog boundaries, settle candidates \\

Combat        & 1 & Damage estimates, modifiers, kill/death predictions \\

Production    & 2 & Available units/buildings/districts with cost and turns \\

Tech / Civics & 1 & Available research with turns, eureka/inspiration status \\

Diplomacy     & 4 & Relationships, modifiers, sessions, pending deals \\

Trade         & 4 & Active routes, destinations with yields, deal composition \\

Governance    & 4 & Policies, governors, unit promotions \\

Religion      & 3 & Pantheon beliefs, founding status, per-city religious spread \\

Espionage     & 1 & Spy positions, ranks, available missions \\

World Congress & 1 & Resolutions, voting options, passed effects \\

Victory       & 1 & Per-type progress, rival intelligence, viability assessment \\

Great People  & 2 & Candidates, recruitment race, activation status \\

Resources     & 1 & Strategic stockpiles, luxury/bonus tiles, nearby unclaimed \\

Notifications & 1 & Game events, turn blockers, required actions \\

\bottomrule

\end{tabular}

\end{table}

\FloatBarrier

\subsection{Narration Output Examples}

\label{sec:gamecode}

\refstepcounter{gamecode}\label{gamecode:overview-narration}

\begin{gamecode}

Turn 87/330 | Scythia (Tomyris) | Score: 198

Gold: 156 (+12/turn) | Science: 24.3 | Culture: 18.7 | Faith: 89 | Favor: 23 (+3/turn)

Research: IRON_WORKING (3 turns) | Civic: MILITARY_TRADITION (5 turns)

Cities: 4 | Population: 14 | Units: 9

Explored: 34% of land (412/1212 tiles)

Religion: none yet (2/4 slots remaining -- Great Prophet needed)

Era: CLASSICAL | Score: 18 (Dark: 12, Golden: 24)

Rankings: Babylon 245 > Scythia 198 > France 187 > Korea 156

\end{gamecode}

\refstepcounter{gamecode}\label{gamecode:unit-narration}

\begin{gamecode}

9 units:

  Saka Horse Archer (UNIT_SCYTHIAN_HORSE_ARCHER) at (11,23)

    -- CS:20 RS:25 moves 2/4 [id:65538, idx:2]

    >> CAN ATTACK: Barbarian WARRIOR at (13,24) -- 2 tiles, ranged

  Builder (UNIT_BUILDER) at (10,22) -- moves 2/2 charges:2 [id:131074, idx:3]

    >> Can build: IMPROVEMENT_MINE, IMPROVEMENT_FARM, IMPROVEMENT_PLANTATION

  Settler (UNIT_SETTLER) at (9,22) -- moves 0/2 (no moves) [id:196610, idx:4]

  Warrior (UNIT_WARRIOR) at (12,23) -- CS:20 moves 2/2

    CAN UPGRADE to Swordsman (120g) [id:65537, idx:1]

Nearby threats (1):

  Barbarian (1 unit):

    WARRIOR at (13,24) -- CS:20 HP:100/100 (3 tiles away)

\end{gamecode}

\refstepcounter{gamecode}\label{gamecode:diplomacy-narration}

\begin{gamecode}

3 civilizations:

  Babylon (Hammurabi) -- FRIENDLY (+12) [player 1]

    Cities (6): Babylon pop 14 (20,15); Sippar pop 8 (18,18) [walls] + 4 in fog

    Military: 245 vs our 156 (1.6x) !! MUCH STRONGER

    Access: we have delegation, we have embassy

    +6 Delegation

    +3 Open Borders

    +3 Shared Agenda: likes civilizations with strong military

  France (Catherine) -- NEUTRAL (+2) [player 2]

    Cities (5): Paris pop 11 (31,9) [walls]; Marseille pop 6 (28,12)

    Military: 198 vs our 156 (1.3x)

    Access: we have delegation

    -3 Different Government

    +5 Delegation

\end{gamecode}

\subsection{Performance Signals}

\begin{itemize}
\label{list:continuous_performance_signals}
\item \textbf{Score trajectory.} Game score decomposed by category at every turn.
\item \textbf{Yield curves.} Gold, science, culture, and faith per turn tracked longitudinally.
\item \textbf{City development.} Population growth, district completion rates, production throughput.
\item \textbf{Military effectiveness.} Kill/loss ratio, territory held, threat response latency.
\item \textbf{Diplomatic standing.} Relationship modifiers, grievance accumulation, suzerainty count.
\item \textbf{Technology pace.} Mean turns per technology, eureka and inspiration capture rates.
\item \textbf{Expansion timing.} Turns to second, third, and fourth city.
\end{itemize}

\subsection{A Turn in Practice}

\label{sec:turnexample}

The following annotated example illustrates a single turn as the agent

experiences it. The agent begins by orienting, identifies a threat, acts on

it, then reflects.

\begin{tcolorbox}[title={\small\textbf{Worked Example: One Complete Turn}}, colback=gray!3, colframe=gray!40]

\small

\textbf{Step 1.} Agent calls \texttt{get\_game\_overview}:

\begin{Verbatim}[fontsize=\footnotesize]

Turn 87/330 | Scythia (Tomyris) | Score: 198

Gold: 156 (+12/turn) | Science: 24.3 | Culture: 18.7 | Faith: 89

Rankings: Babylon 245 > Scythia 198 > France 187 > Korea 156

\end{Verbatim}

\textbf{Step 2.} Agent calls \texttt{get\_units} and sees a threat:

\begin{Verbatim}[fontsize=\footnotesize]

Saka Horse Archer at (11,23) -- CS:20 RS:25 moves 4/4

  >> CAN ATTACK: Barbarian WARRIOR at (13,24) -- 2 tiles, ranged

Nearby threats (1):

  Barbarian (1 unit): WARRIOR at (13,24) -- CS:20 HP:100/100 (3 tiles away)

\end{Verbatim}

\textbf{Step 3.} Agent reasons: \emph{"Barbarian warrior 3 tiles from my undefended settler. My Saka Horse Archer has range 2 and 4 moves---I can intercept with a ranged attack without taking damage."}

\textbf{Step 4.} Agent calls

\texttt{unit\_action(unit\_id=65538, action=`attack', target\_x=13, target\_y=24)}:

\begin{Verbatim}[fontsize=\footnotesize]

Combat Estimate (Ranged):

  SCYTHIAN_HORSE_ARCHER (CS:20, HP:100) vs WARRIOR (CS:20, HP:100)

  Est damage to defender: ~45

  Attacked Barbarian WARRIOR at (13,24) -- dealt 47 damage (HP: 100 -> 53)

\end{Verbatim}

\textbf{Step 5.} After handling remaining units and production, agent calls

\texttt{end\_turn} with diary:

\begin{Verbatim}[fontsize=\footnotesize]

tactical:   "Intercepted barbarian warrior threatening settler. 47 damage

             via ranged, will finish next turn. Builder improved iron at (11,22)."

strategic:  "Babylon leads by 47 score. Need 3rd city soon -- settler ready

             next turn."

planning:   "T88: finish barbarian, move settler east. T89: found city at (15,25)."

hypothesis: "Babylon likely to declare friendship -- positive modifiers trending up."

tooling:    "No issues."

\end{Verbatim}

\end{tcolorbox}

\subsection{Narration and Playbook Design Principles}

\begin{enumerate}

\label{list:design_principles}

\item \textbf{Flag urgency.} Threats are bold-marked (\texttt{**[Barbarian WARRIOR]**}), unimproved resources receive \texttt{!!} warnings, and critical states such as loyalty crises or starvation are explicitly highlighted. The narration is intentionally selective about what is emphasised.

\item \textbf{Provide context for action.} Unit readouts include valid attack targets and buildable improvements. City readouts include available production and defensive status. The agent is presented with actionable options rather than raw state alone.

\item \textbf{Compress intelligently.} Fog-of-war tiles are marked \texttt{[fog]} rather than omitted, so the agent can distinguish between absence of information and absence of content. Resources are classified by type (bonus/luxury\textsuperscript{+}/strategic\textsuperscript{*}) to support prioritisation, and rankings are sorted by score.

\end{enumerate}

\paragraph{Strategic Guidance with Benchmarks.}

The playbook uses advisory rather than imperative language. Earlier versions included hard IF/WHEN triggers (e.g.\ "IF gold > 500: spend before ending turn"), but these were softened to allow more organic behaviour. The agent frequently violates even this guidance, making the reflection-action gap observable against instructions it demonstrably interprets but does not consistently follow.

The playbook therefore serves both as a stabilisation mechanism and as an experimental control. It standardises interaction structure and highlights relevant signals while leaving attention allocation and execution decisions to the agent.

This cycle of orienting, detecting threats, reasoning, acting, and reflecting repeats across hundreds of game turns, producing a structured trajectory of situated, tool-mediated decision-making.

\subsection{Agent Playbook Excerpt}

\label{app:playbook}

The following excerpt from the agent playbook (200+ lines, iteratively refined

across ten games, available in full at the repository) illustrates three design

elements: sensorium awareness framing, advisory strategic checkpoints, and

the structured turn diary.

\paragraph{Sensorium Awareness}

\begin{quote}

\textbf{You only know what you explicitly query.} A human player passively

absorbs the minimap, score ticker, religion lens, unit health bars---you have

none of that. Information you don't ask for simply doesn't enter your world

model. The checkpoints and patterns below exist to compensate for this.

\end{quote}

\begin{quote}

\textbf{Gold.} Gold sitting above 500 with no specific plan is usually better

deployed. A builder, a luxury tile, a building that skips 5+ turns of

production---these compound. Saving for a specific purchase is fine, but it

helps to name the item and the turn.

\textbf{Expansion.} Each city multiplies your districts, yields, and Great

Person generation. The gap between a 3-city and 5-city empire at T100 is hard

to recover from. Benchmarks: T40: 2 cities, T60: 3 cities, T80: 4 cities,

T100: 4--5 cities. If city count is lagging, a settler is typically the

highest-impact production choice.

\textbf{Exploration.} You can't settle what you can't see, and you can't

counter threats you don't know exist. A scout set to \texttt{automate} is one

of the best investments in the early game. Benchmarks: T25 $\geq$15\%, T50

$\geq$25\%, T75 $\geq$35\%, T100 $\geq$50\%.

\end{quote}

\paragraph{Strategic Checkpoints (Every 20 Turns).}

\begin{quote}

\begin{itemize}

\item \texttt{get\_diplomacy}---delegations to new civs, friendships with

Friendly civs, alliances if eligible.

\item \texttt{get\_victory\_progress}---check all 6 victory types, not just

your own path.

\item \texttt{get\_religion\_spread}---religious victory is invisible without

active checking; a rival with majority in most civs is a serious threat.

\end{itemize}

\end{quote}

\paragraph{Turn Diary Fields.}

\begin{quote}

Five fields each turn:

\begin{itemize}

\item \textbf{tactical}: What happened---specific units, tiles, outcomes.

\item \textbf{strategic}: Standings vs rivals---yields, city count, victory

path viability with numbers.

\item \textbf{tooling}: Tool issues observed, or ``No issues''.

\item \textbf{planning}: Concrete actions for the next 5--10 turns---specific

builds, moves, research targets with turn estimates.

\item \textbf{hypothesis}: Specific predictions---attack timing, milestone

turns, biggest risks.

\end{itemize}

\end{quote}

\subsection{Inspect Framework}

\label{sec:inspect}

An open-source evaluation platform with native MCP support. Each game scenario maps to an Inspect \texttt{Sample}, the MCP server is consumed directly via \texttt{mcp\_server\_stdio()}, and a custom scorer extracts per-dimension metrics from tool-call transcripts. Inspect provides model portability (15+ providers), structured logging, and checkpoint recovery---important for multi-hour game sessions.

\FloatBarrier
\section{Supplementary Results}

\label{app:supplementary}

% Placeholder for supplementary tables and figures.

\subsection{Scores}

\label{sec:scores}

\begin{figure}[htbp]

\centering

\includegraphics[width=\textwidth]{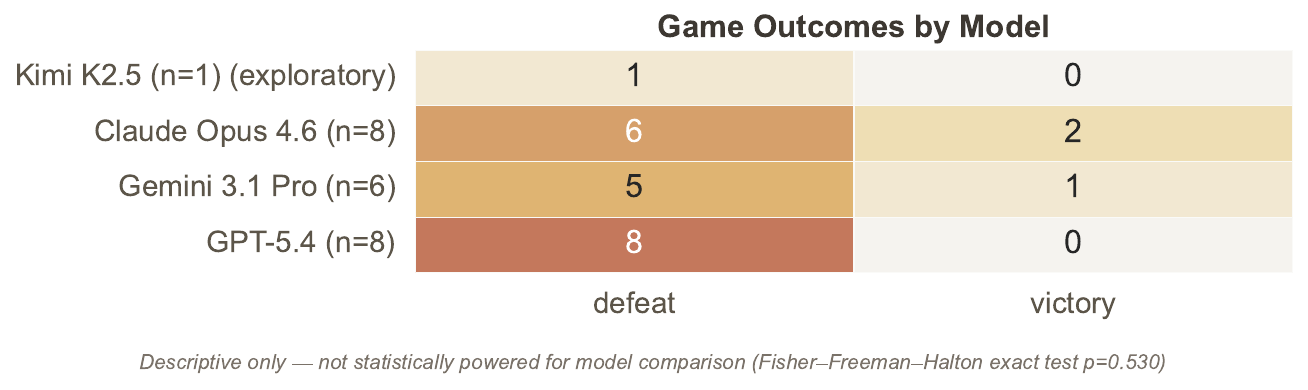}

\caption{\textbf{Outcome heatmap}}

\label{fig:outcome_heatmap}

\end{figure}

\begin{figure}[htbp]

\centering

\includegraphics[width=\textwidth]{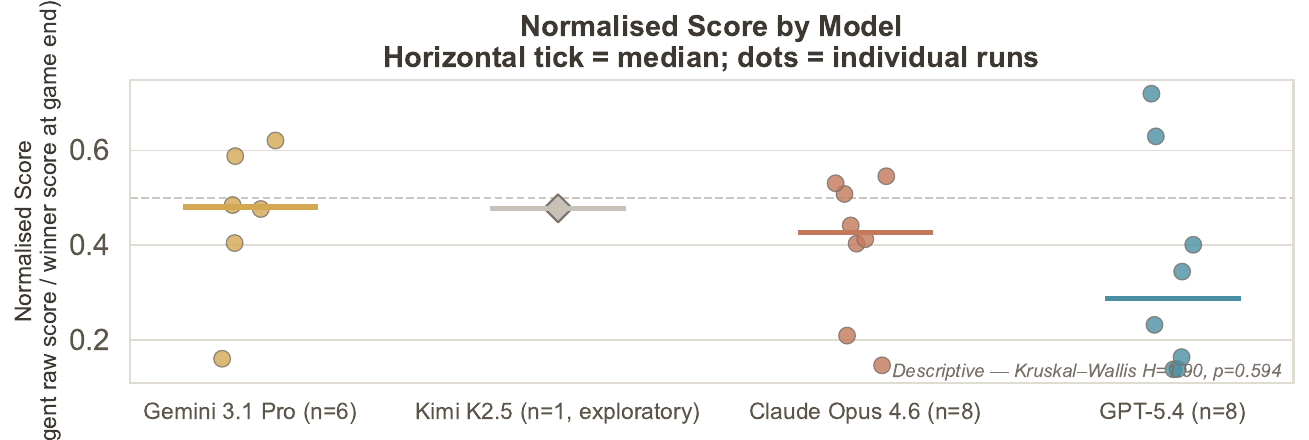}

\caption{\textbf{Normalised score} (agent raw score / winner raw score at game end) by model across all games.}

\label{fig:normalised_score}

\end{figure}

\begin{figure}[htbp]

\centering

\includegraphics[width=\textwidth]{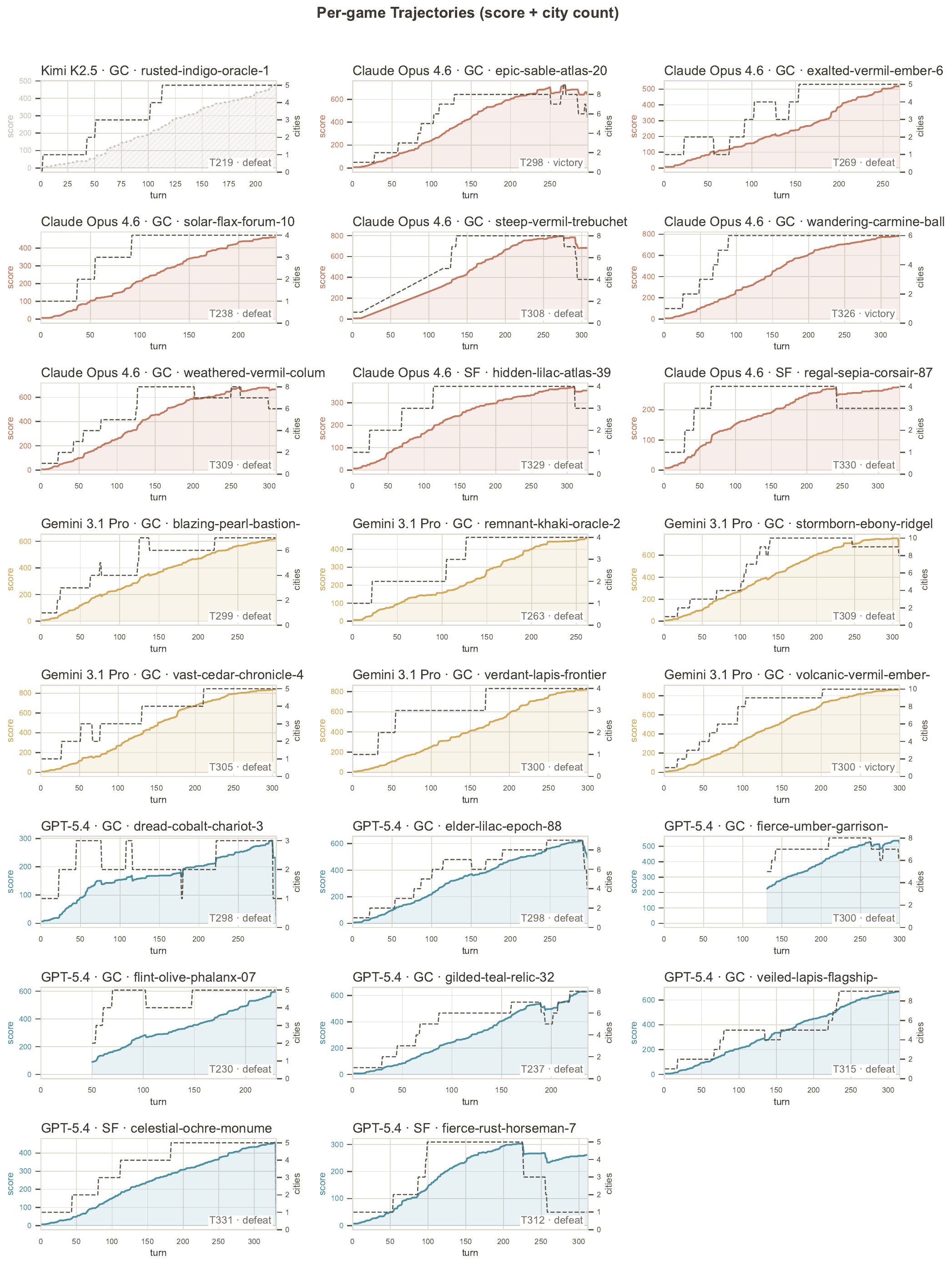}

\caption{\textbf{Per-game trajectory overview} for all games.}

\label{fig:sparklines}

\end{figure}

\begin{figure}[htbp]

\centering

\includegraphics[width=\textwidth]{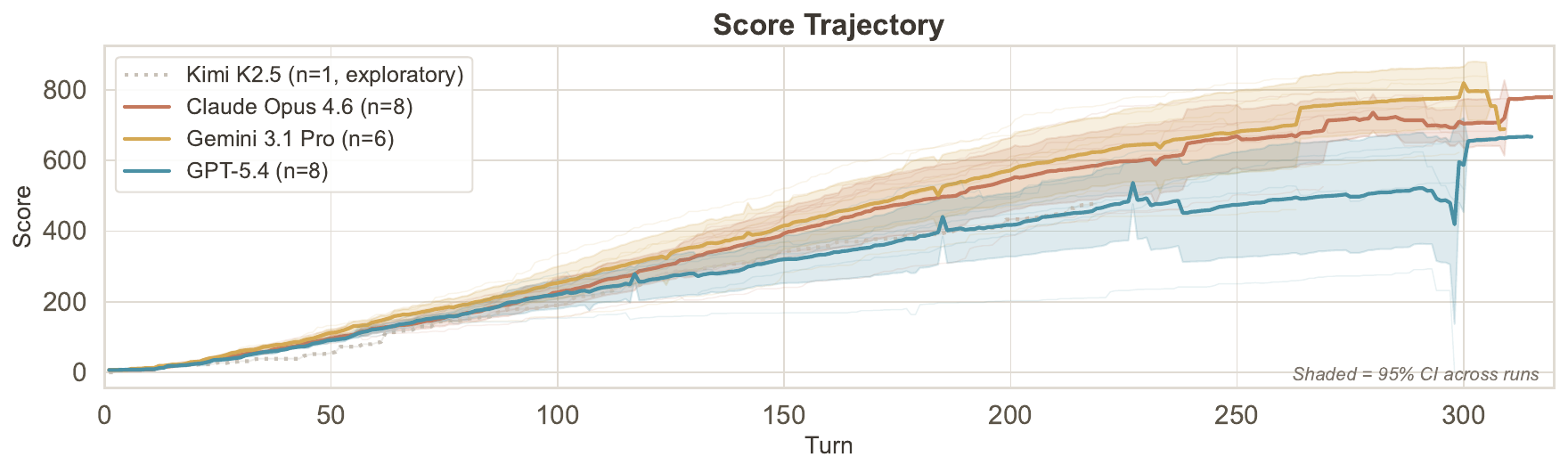}

\caption{\textbf{Raw score trajectories.} The shaded band shows 95\% CI.}

\label{fig:trajectory_score_ground_control}

\end{figure}

\begin{figure}[htbp]

\centering

\includegraphics[width=\textwidth]{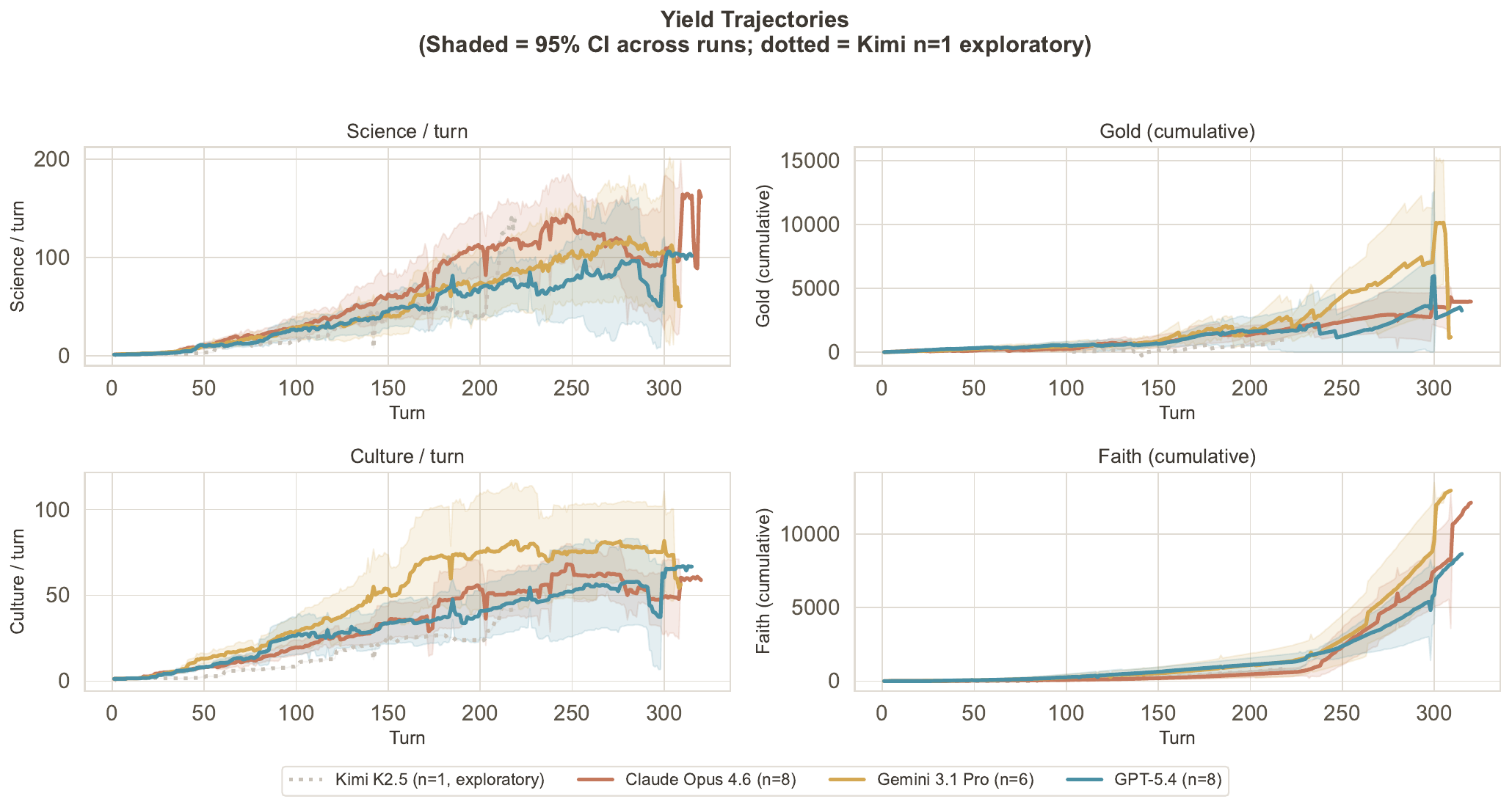}

\caption{\textbf{Per-turn yield trajectories.} The shaded band 95\% CI.}

\label{fig:yield_trajectories_ground_control}

\end{figure}

\begin{table}[htbp]

  \centering

  \caption{Outcome breakdown by model and scenario.}

  \label{tab:model-scenario}

  \begin{tabular}{llrrr}

    \toprule

    Model & Scenario & Games & Victories & Defeats \\

    \midrule

    Claude Opus 4.6 & Ground Control & 6 & 2 & 4 \\

    Claude Opus 4.6 & Snowflake & 2 & 0 & 2 \\

    Gemini 3.1 Pro & Ground Control & 6 & 1 & 5 \\

    GPT-5.4 & Ground Control & 6 & 0 & 6 \\

    GPT-5.4 & Snowflake & 2 & 0 & 2 \\

    Kimi-K2.5 & Ground Control & 1 & 0 & 1 \\

    \bottomrule

  \end{tabular}

\end{table}

\FloatBarrier

\subsection{Statistical Power}

\begin{table}[htbp]

\centering

\caption{Discriminative Power (ICC) --- \texttt{ground\_control} ICC estimates are exploratory because the number of runs per model is small and uneven.}

\label{tab:icc_table}

\begin{tabular}{lrrrrr}

\toprule

Metric & ICC & Within SD & Between SD & Test & Verdict \\

\midrule

exploration\_t100 & 0.717 & 3.628 & 5.515 & H=12.28, p=0.0065 & \textbf{exploratory signal} \\

exploration\_t50 & 0.458 & 4.401 & 4.524 & H=11.08, p=0.0113 & marginal \\

turns\_played & 0.197 & 28.396 & 35.668 & H=3.87, p=0.2758 & noise \\

final\_exploration\_pct & 0.111 & 34.636 & 28.099 & H=4.25, p=0.2359 & noise \\

final\_science & 0.039 & 47.149 & 38.812 & H=2.23, p=0.5269 & noise \\

final\_culture & 0.029 & 27.520 & 12.484 & H=2.33, p=0.5076 & noise \\

final\_military & -0.035 & 416.790 & 191.236 & H=2.90, p=0.4073 & noise \\

final\_gold & -0.078 & 2947.950 & 1532.178 & H=1.89, p=0.5957 & noise \\

city\_t100 & -0.124 & 1.499 & 0.788 & H=3.34, p=0.3427 & noise \\

final\_cities & -0.170 & 2.174 & 0.631 & H=0.70, p=0.8722 & noise \\

city\_t200 & -0.170 & 2.180 & 0.701 & H=0.86, p=0.8351 & noise \\

city\_t150 & -0.185 & 2.268 & 0.644 & H=0.81, p=0.8459 & noise \\

city\_t50 & -0.221 & 1.052 & 0.397 & H=0.83, p=0.8425 & noise \\

\bottomrule

\end{tabular}

\end{table}

\FloatBarrier

\subsection{Milestones and Progression}
\begin{figure}[htbp]

\centering

\includegraphics[width=\textwidth]{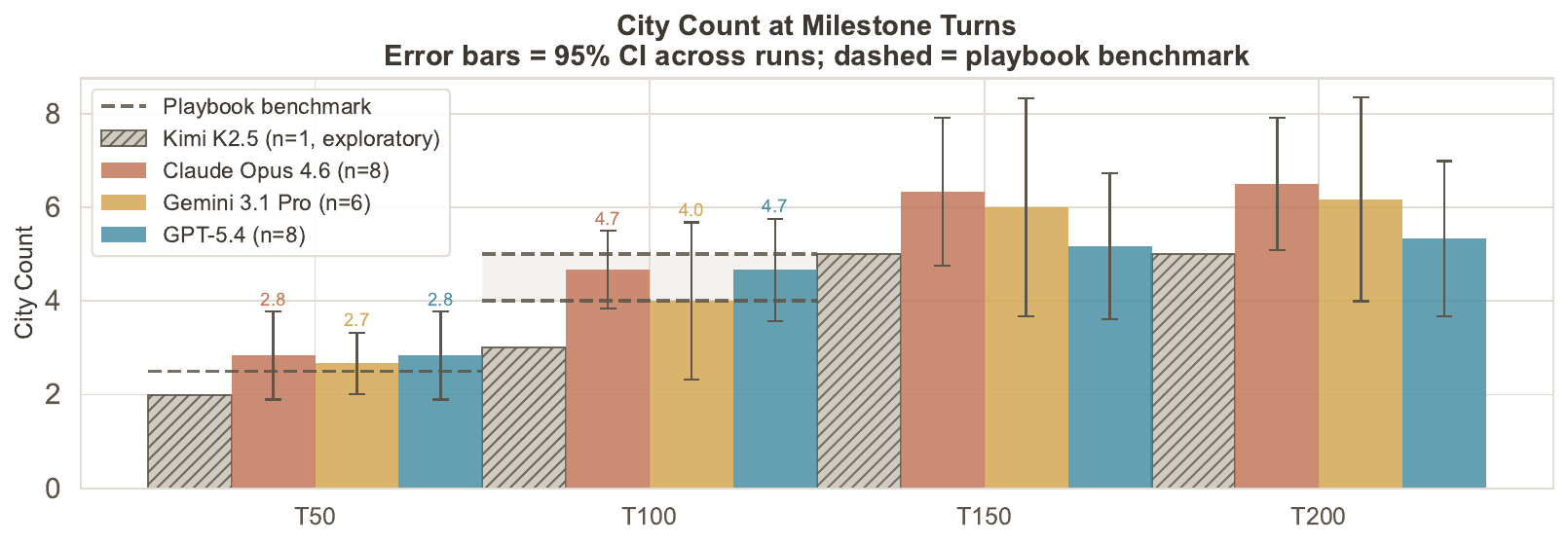}

\caption{\textbf{Mean city count at turn checkpoints.}}

\label{fig:city_milestones_ground_control}

\end{figure}

Early expansion is a useful trajectory-level signal because additional cities compound over time. At T50, agents averaged \CityTFiftyClaude{} cities for Claude, \CityTFiftyGPT{} for GPT, \CityTFiftyGemini{} for Gemini, and \CityTFiftyKimi{} for Kimi (Fig.~\ref{fig:city_milestones_ground_control}). By T100, the corresponding averages were \CityTHundredClaude{}, \CityTHundredGPT{}, \CityTHundredGemini{}, and \CityTHundredKimi{}. Most models therefore reached the playbook's T100 target of 4--5 cities on average, although individual trajectories varied substantially. These results illustrate how CivBench captures variation in expansion timing and development across runs.

\begin{figure}[htbp]

\centering

\includegraphics[width=\textwidth]{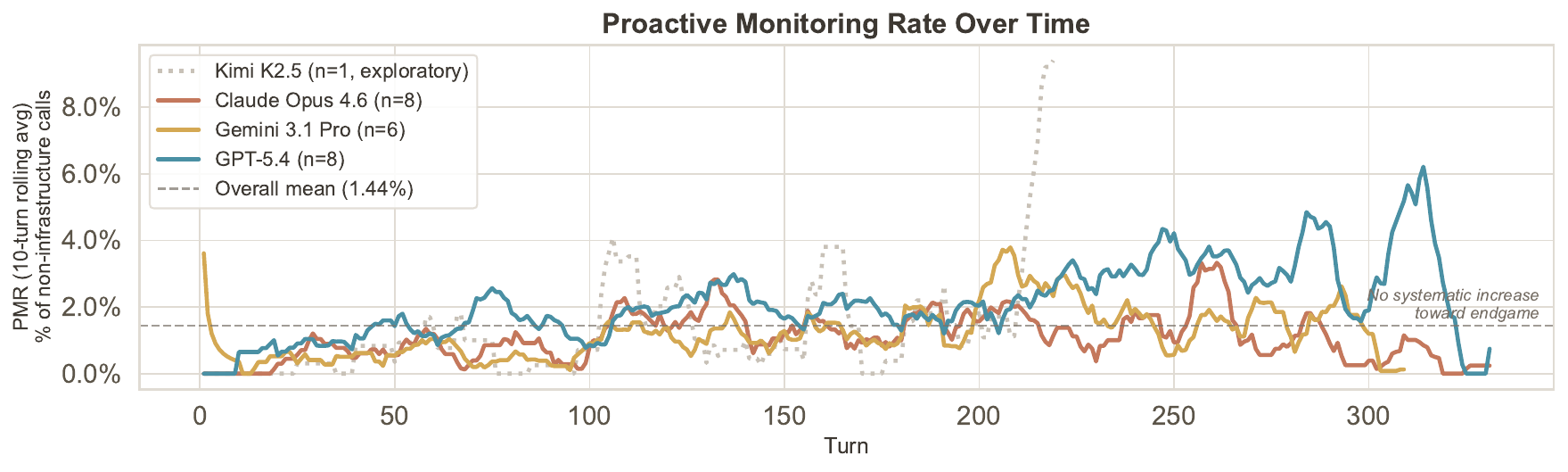}

\caption{\textbf{Proactive Monitoring Rate} by game turn (10-turn rolling window) per model.}

\label{fig:pmr_over_time}

\end{figure}

\FloatBarrier

\subsection{Tools}

\begin{figure}[htbp]

\centering

\includegraphics[width=\textwidth]{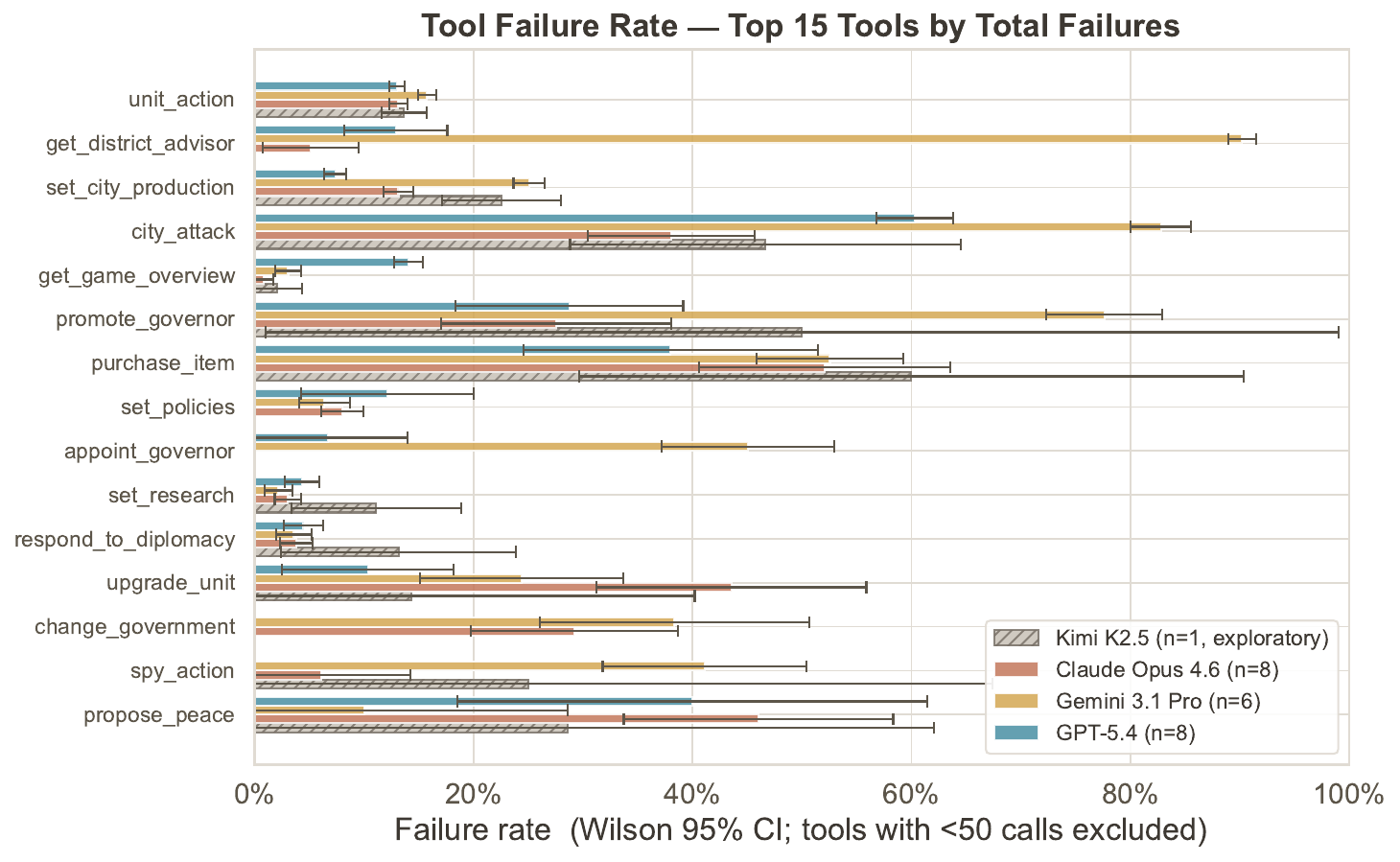}

\caption{\textbf{Tool failure rate} per tool and model.}

\label{fig:tool_error_rate}

\end{figure}

\begin{figure}[htbp]

\centering

\includegraphics[width=\textwidth]{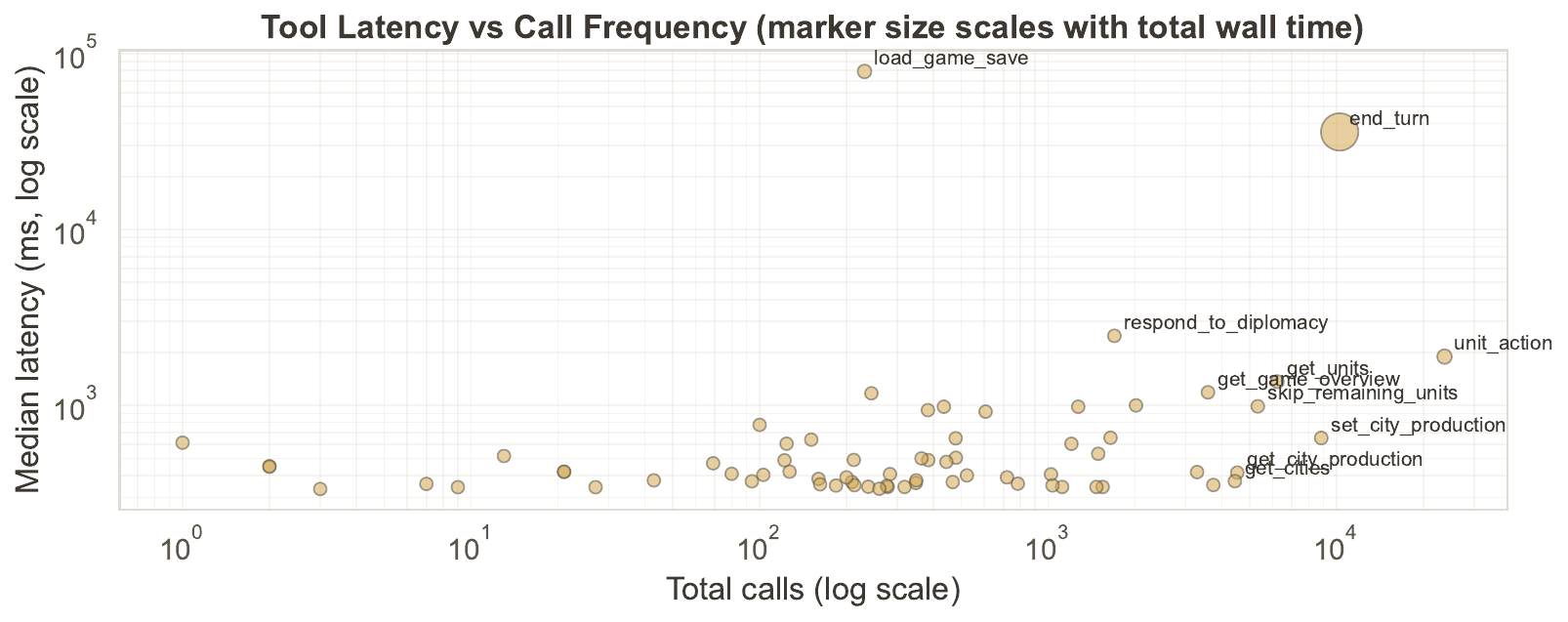}

\caption{Median round-trip duration (ms) vs total call count per tool.}

\label{fig:tool_latency}

\end{figure}

\begin{figure}[htbp]

\centering

\includegraphics[width=\textwidth]{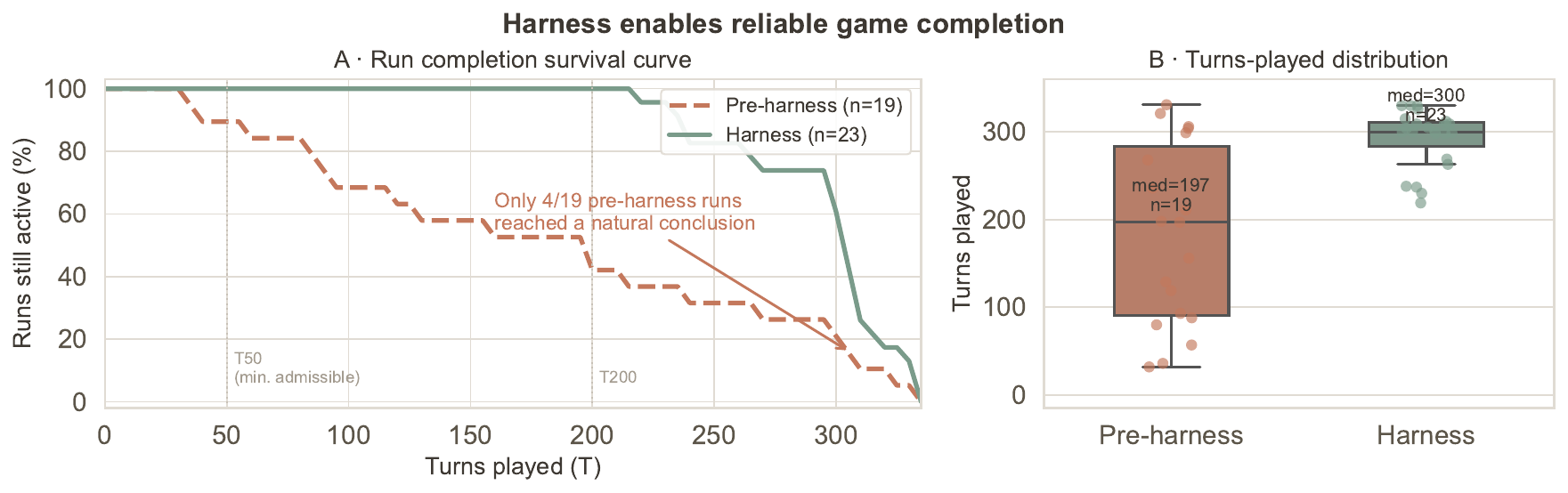}

\caption{Left: Harness vs pre-harness turn rates. Right: Turns played with the harness vs pre-Harness. Pre-harness runs used the most basic possible infrastructure to run, are not admissible and frequently did not finish games.}

\label{fig:harness_comparison}

\end{figure}

\FloatBarrier

\subsection{Yield Trajectories}

\label{app:trajectories}

% Gold/turn, science/turn, culture/turn averaged across 5 runs per model.

% Line plot with shaded variance bands.

% Rising gold/turn with high peak hoard = continuous-signal sensorium

% effect correlate.

Score trajectories diverge early and amplify through the midgame (Fig~\ref{fig:trajectory_score_ground_control}). All models grow at similar rates through approximately T50, after which developmental arcs separate: the highest-trajectory runs open score leads that persist to game end, while the lowest accumulate deficits that are not recoverable within the game horizon. This divergence does not map proportionally onto normalised score, for the reason given in Section~\ref{sec:results-aggregate} --- raw score describes a model's absolute developmental arc, normalised score its standing relative to whoever actually won. A Claude game that ran to T326 with a strong Korea opponent can produce a lower normalised score than a GPT game terminated at T230 against a weaker opponent set. At T50 raw scores are closely bunched (Claude $\TFiftyScoreClaude{} \pm \TFiftypmScoreClaude{}$, Gemini $\TFiftyScoreGemini{} \pm \TFiftypmScoreGemini{}$, GPT $\TFiftyScoreGPT{} \pm \TFiftypmScoreGPT{}$) but by T150 Claude ($\THundredFiftyScoreClaude{} \pm \THundredFiftypmScoreClaude{}$) and Gemini ($\THundredFiftyScoreGemini{} \pm \THundredFiftypmScoreGemini{}$) substantially lead GPT ($\THundredFiftyScoreGPT{} \pm \THundredFiftypmScoreGPT{}$), and the gap persists through T200 (Claude $\TTwoHundredScoreClaude{} \pm \TTwoHundredpmScoreClaude{}$ vs GPT $\TTwoHundredScoreGPT{} \pm \TTwoHundredpmScoreGPT{}$), which persists until the end of the game.

Per-turn yield decomposition (Fig~\ref{fig:yield_trajectories_ground_control}) shows that score differences are driven primarily by science output, which exhibits the clearest and most sustained between-model separation. Culture trajectories cross repeatedly and track final culture's low discriminative power in Table~\ref{tab:icc_table}. Cumulative gold and faith are dominated by game-length effects rather than per-turn efficiency differences. The pattern is consistent with Civilization~VI's underlying mechanics: science compounds through technology unlocks that enable further yield improvements, while culture and faith contribute to score more directly but with weaker feedback on subsequent yields. We are not aware of comparable per-turn economic trajectories reported across LLM agents in a commercial strategy game, allowing us to observe when behaviour diverges.

\begin{table}[t]
\centering
\caption{\textbf{Sensitivity of RAG to the evaluation horizon.} Execution fidelity for a fixed commitment set evaluated over 5, 10, and 20 subsequent turns; higher values indicate greater execution fidelity.}
\label{tab:rag_horizon_sensitivity}
\small
\begin{tabular}{lrrr}
\toprule
\textbf{Model} & \textbf{RAG@5} & \textbf{RAG@10} & \textbf{RAG@20} \\
\midrule
Claude Opus 4.6 & \RAGFiveClaude{}\% & \RAGClaude{}\% & \RAGTwentyClaude{}\% \\
Gemini 3.1 Pro & \RAGFiveGemini{}\% & \RAGGemini{}\% & \RAGTwentyGemini{}\% \\
GPT-5.4 & \RAGFiveGPT{}\% & \RAGGPTFive{}\% & \RAGTwentyGPT{}\% \\
\midrule
\textbf{Overall (pooled)} & \textbf{\RAGPooledFive{}\%} & \textbf{\RAGPooledTen{}\%} & \textbf{\RAGPooledTwenty{}\%} \\
\bottomrule
\end{tabular}
\end{table}

\subsection{RAG Horizon Sensitivity}
\label{sec:rag_sensitivity}
We examine the sensitivity of RAG to the evaluation horizon using $K=5$, $K=10$, and $K=20$ (Table~\ref{tab:rag_horizon_sensitivity}). The commitment set is held fixed, and execution is evaluated using the same executed/partial/not-executed rubric at each horizon. The $K=5$ and $K=20$ labels were derived with reference to the $K=10$ classifications, using automated propagation, LLM-assisted assessment, and three manual corrections. RAG increases as the evaluation window widens across all three models, while their ordering remains unchanged. The pooled RAG score increases from \RAGPooledFive{}\% at $K=5$ to \RAGPooledTen{}\% at $K=10$ and \RAGPooledTwenty{}\% at $K=20$. Even at 20 turns, \RAGTwentyNotExecuted{}\% of commitments remain unexecuted and a further \RAGTwentyPartial{}\% are only partially executed. The finding that agents frequently fail to execute stated commitments therefore persists across the tested horizons.

\FloatBarrier

\subsection{RAG Sampling}
\label{sec:rag_sampling}
To examine commitment execution across different stages of gameplay, we selected diary turns covering broadly early, middle, and late game. The sampling plan comprised 35 turns across 14 runs. One turn contained no planning text, while seven further turns (six GPT-5.4 and one Claude) could not be processed due to unparseable labeller outputs. The resulting analysis comprises 142 commitments from 27 turns across 13 runs. As the sample was not randomly selected, the reported RAG results are descriptive of the analysed commitments.

\FloatBarrier
\section{Feature Comparison}

Table~\ref{tab:civrealm} summarises the main differences between CivRealm and CivBench, with particular emphasis on the game engine, interface, and strategic mechanics relevant to evaluation.

\begin{table}[htbp]
\centering
\caption{Feature comparison: CivRealm (FreeCiv) vs.\ CivBench (Civilization~VI).}
\label{tab:civrealm}
\small
\begin{tabular}{lll}
\toprule
\textbf{Feature} & \textbf{CivRealm (FreeCiv)} & \textbf{CivBench (Civ VI)} \\
\midrule
Game engine        & FreeCiv (open-source, Civ II-era) & Civilization VI (commercial, 2016+) \\
Grid type          & Square, 8-directional             & Hex, 6-directional \\
Unit stacking      & Allowed                           & One-unit-per-tile \\
Districts          & No                                & Yes (13 types + unique) \\
Governors          & No                                & Yes (7 governors, promotions) \\
Loyalty/amenities  & No                                & Yes (city loyalty, amenity system) \\
Espionage          & Basic                             & Full (missions, counterintel) \\
World Congress     & No                                & Yes (resolutions, diplomatic VP) \\
Great People       & Basic                             & Full (recruitment race, activation) \\
Government         & Tax sliders (sci/tax/lux)         & Policy card slots + government types \\
Religion           & Basic                             & Full (pantheon, found, spread, inquisition) \\
Victory conditions & 3                                 & 6 \\
Agent interface    & Gymnasium tensor / LangChain text & MCP tool calls + narration layer \\
Research Items       & 87                          & 117 \\
LLM training data  & Minimal (niche open-source)       & Extensive (a decade of guides, forums) \\
\bottomrule
\end{tabular}
\end{table}

\begin{table}[t]

\centering

\caption{Evaluation dimensions.}

\label{tab:dimensions}

\small

\begin{tabular}{lll}

\toprule

\textbf{Dimension} & \textbf{Signal} & \textbf{Measurement} \\

\midrule

Overall Score          & Aggregate score         & Civ~6 score at checkpoints \\

Economic Management    & Yield optimisation      & Gold, science, culture per turn \\

Military Competence    & Threat response         & Attack actions, kill/loss ratio \\

Scientific Progress    & Technology pace         & Turns per technology, eureka rate \\

Diplomatic Skill       & Relationship management & Diplomatic actions, favour income \\

Spatial Reasoning      & Map awareness           & Exploration, cities founded \\

Tool-Use Fluency       & Tool selection          & Error rate, tool diversity \\

Long-Horizon Coherence & Strategy consistency    & Score variance, drift \\

\bottomrule

\end{tabular}

\end{table}

\begin{figure}[t]

\centering

\includegraphics[width=\textwidth]{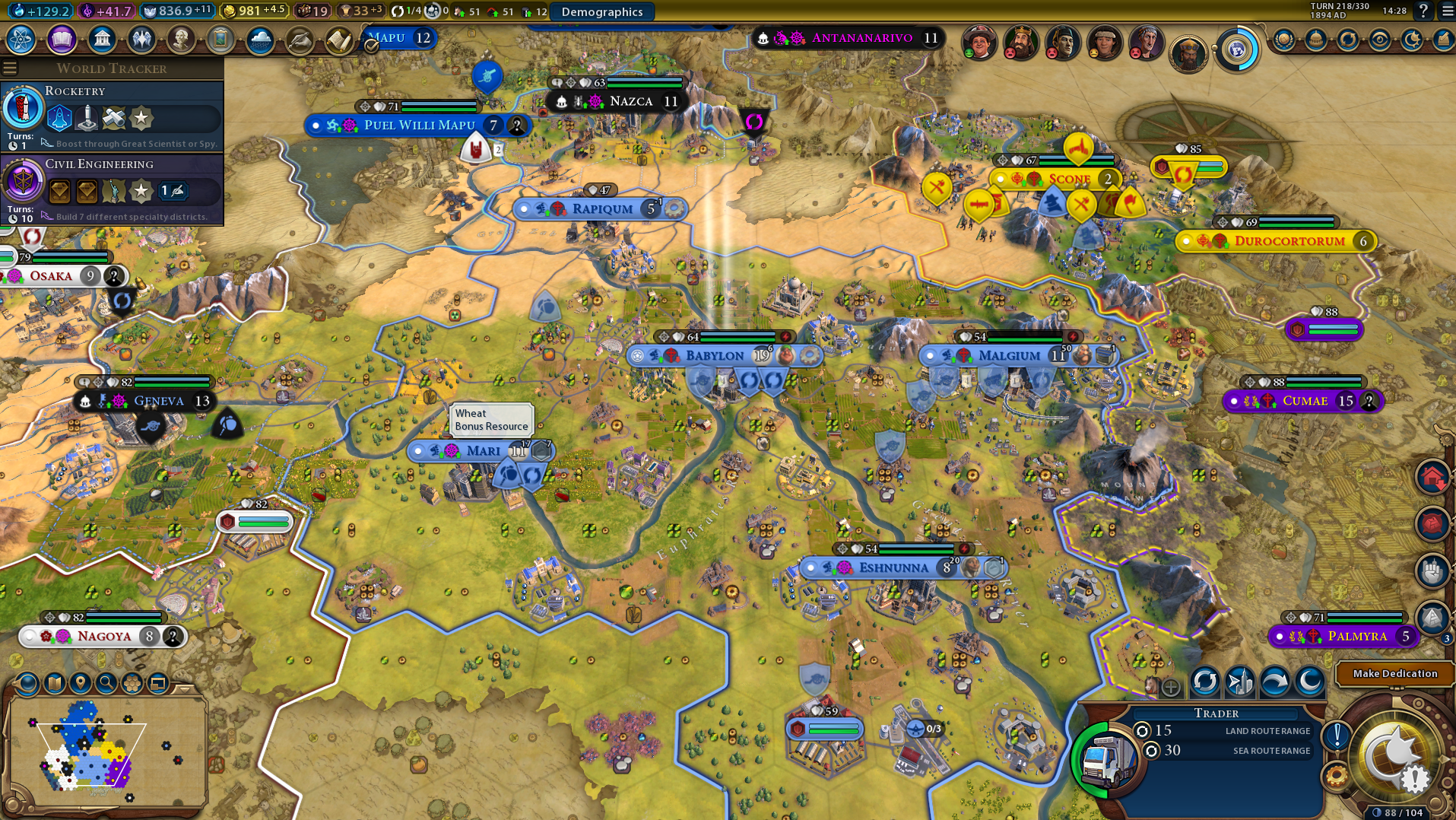}

\caption{\textbf{Screenshot of a Civilization~VI game.} This illustrates what a human player would see.}

\label{fig:game_screenshot}

\end{figure}

\FloatBarrier
\section{Tool Inventory}

This appendix section lists the tool categories used by CivBench and the representative MCP calls in each category.

\begin{table}[htbp]
\centering
\caption{Tool inventory by category.}
\label{tab:tools}
\small
\begin{tabular}{llp{7cm}}
\toprule
\textbf{Category} & \textbf{Count} & \textbf{Representative tools} \\
\midrule
State queries    & 27 & \texttt{get\_game\_overview}, \texttt{get\_units},
                        \texttt{get\_map\_area}, \texttt{get\_diplomacy},
                        \texttt{get\_victory\_progress} \\
Unit actions     & 10 & \texttt{unit\_action}, \texttt{upgrade\_unit},
                        \texttt{promote\_unit}, \texttt{skip\_remaining\_units} \\
City management  & 9  & \texttt{set\_city\_production}, \texttt{purchase\_item},
                        \texttt{purchase\_tile}, \texttt{set\_city\_focus} \\
Diplomacy        & 7  & \texttt{respond\_to\_diplomacy}, \texttt{respond\_to\_trade},
                        \texttt{send\_diplomatic\_action}, \texttt{propose\_trade},
                        \texttt{form\_alliance} \\
Governance       & 10 & \texttt{set\_research}, \texttt{set\_policies},
                        \texttt{change\_government}, \texttt{appoint\_governor} \\
Religion \& culture & 5 & \texttt{choose\_pantheon}, \texttt{choose\_dedication},
                          \texttt{vote\_world\_congress} \\
Game lifecycle   & 8  & \texttt{end\_turn}, \texttt{get\_diary},
                        \texttt{dismiss\_popup}, \texttt{quicksave},
                        \texttt{load\_save}, \texttt{restart\_and\_load} \\
\midrule
\textbf{Total} & \textbf{76} & \\
\bottomrule
\end{tabular}
\end{table}

\FloatBarrier
\newpage
\section{Model Access Dates}
\label{sec:models_access}

\begin{table}[htbp]
  \centering
  \small
  \caption{Model identifiers and access window summary (admissible runs only, $N=\NTotalGames{}$). Inspect AI model id is \texttt{provider/host/model}; all OpenAI and Kimi models routed through Azure used the Chat Completions API (Azure lacks full Responses API parity at access time).}
  \label{tab:model-summary}
  \begin{tabular}{llllr}
    \toprule
    Paper label & Inspect AI model id & Endpoint & Access window (UTC) & Runs \\
    \midrule
    Claude Opus 4.6        & \texttt{anthropic/vertex/claude-opus-4-6}        & GCP Vertex AI     & 2026-04-02 -- 2026-04-16 & 8 \\
    GPT-5.4                & \texttt{openai/azure/gpt-5.4}                    & Azure OpenAI      & 2026-04-05 -- 2026-04-16 & 8 \\
    Gemini 3.1 Pro & \texttt{google/vertex/gemini-3.1-pro-preview}    & GCP Vertex AI     & 2026-04-12 -- 2026-04-17 & 6 \\
    Kimi-K2.5              & \texttt{openai/azure/Kimi-K2.5}                  & Azure AI Foundry  & 2026-04-18 -- 2026-04-18 & 1 \\
    \bottomrule
  \end{tabular}
\end{table}

\subsection{Reproducibility Details}
\label{sec:reproducibility}

The released code artifact contains all configuration details (Game versions, DLC List, Save files, seeds), and scripts to recompute all metrics and regenerate all figures from the raw tool-call logs.

\paragraph{Artifact contents.}
The released package contains: (1) the CivBench MCP server; (2) scenario definitions and save-file
metadata; (3) raw admissible-game logs; (4) processed metric tables; (5) plotting and analysis
scripts; and (6) documentation for reproducing the reported results. It does not contain
Civilization~VI assets, game binaries, or proprietary Firaxis/2K code.

\paragraph{Known reproducibility risks.}
Reproduction may be affected by Civilization~VI patches, DLC differences, operating-system
differences, provider-side model updates, and stochastic model sampling. We mitigate these risks by
pinning scenario seeds, reporting exact model access dates and settings, logging all tool calls, and
releasing the analysis pipeline.

\FloatBarrier

\newpage
\section*{NeurIPS Paper Checklist}

\begin{enumerate}

\item {\bf Claims}
    \item[] Question: Do the main claims made in the abstract and introduction accurately reflect the paper's contributions and scope?
    \item[] Answer: \answerYes{} % Replace by \answerYes{}, \answerNo{}, or \answerNA{}.
    \item[] Justification: All claims in the abstract and conclusion are supported by the experimental results in Sections 5–6. Scope and power limitations are stated in Section 4.

\item {\bf Limitations}
    \item[] Question: Does the paper discuss the limitations of the work performed by the authors?
    \item[] Answer: \answerYes{} % Replace by \answerYes{}, \answerNo{}, or \answerNA{}.
    \item[] Justification: Section 4 covers commercial dependency, cost, playbook confound, sample size, infrastructure exclusions, non-determinism, and contamination.

\item {\bf Theory assumptions and proofs}
    \item[] Question: For each theoretical result, does the paper provide the full set of assumptions and a complete (and correct) proof?
    \item[] Answer: \answerNA{} % Replace by \answerYes{}, \answerNo{}, or \answerNA{}.
    \item[] Justification: No theoretical results are claimed.

    \item {\bf Experimental result reproducibility}
    \item[] Question: Does the paper fully disclose all the information needed to reproduce the main experimental results of the paper to the extent that it affects the main claims and/or conclusions of the paper (regardless of whether the code and data are provided or not)?
    \item[] Answer: \answerYes{} % Replace by \answerYes{}, \answerNo{}, or \answerNA{}.
    \item[] Justification: The MCP server, dataset, and analysis scripts are open-source at (https://github.com/lmwilki/civ6-mcp, https://huggingface.co/datasets/civbench/civbench-v1/tree/main) Scenario save files are distributed with the repository. 

\item {\bf Open access to data and code}
    \item[] Question: Does the paper provide open access to the data and code, with sufficient instructions to faithfully reproduce the main experimental results, as described in supplemental material?
    \item[] Answer: \answerYes{} % Replace by \answerYes{}, \answerNo{}, or \answerNA{}.
    \item[] Justification: Code at GitHub (MIT license). Game log data are available on HuggingFace (https://github.com/lmwilki/civ6-mcp, https://huggingface.co/datasets/civbench/civbench-v1/tree/main).

\item {\bf Experimental setting/details}
    \item[] Question: Does the paper specify all the training and test details (e.g., data splits, hyperparameters, how they were chosen, type of optimizer) necessary to understand the results?
    \item[] Answer: \answerYes{} % Replace by \answerYes{}, \answerNo{}, or \answerNA{}.
    \item[] Justification: Section 4.1 specifies scenario configurations (map seeds, difficulty, speed), while Appendix E reports model identifiers, endpoints, access dates, and harness versions. The artifact includes full configuration files, playbook versions, and scripts to reproduce all runs and metrics.

\item {\bf Experiment statistical significance}
    \item[] Question: Does the paper report error bars suitably and correctly defined or other appropriate information about the statistical significance of the experiments?
    \item[] Answer: \answerYes{} % Replace by \answerYes{}, \answerNo{}, or \answerNA{}.
    \item[] Justification: Win/loss outcomes are assessed using the Fisher--Freeman--Halton exact test and normalised scores using the Kruskal--Wallis test (Section~\ref{sec:results-aggregate}). PMR results are presented descriptively (Section~\ref{sec:sensorium}). RAG@10 includes 95\% bootstrap confidence intervals and human-label validation (Section~\ref{sec:ragap}), while exploratory ICC results are reported in Appendix~B. Given the small and uneven sample, between-model differences are not interpreted as reliable rankings.

\item {\bf Experiments compute resources}
    \item[] Question: For each experiment, does the paper provide sufficient information on the computer resources (type of compute workers, memory, time of execution) needed to reproduce the experiments?
    \item[] Answer: \answerYes{} % Replace by \answerYes{}, \answerNo{}, or \answerNA{}.
    \item[] Justification: Each game costs approximately \$31--229 in API fees and runs 2--8 hours on a consumer Mac with Civilization~VI running locally. Analysis scripts reproduce all figures in under 5 minutes on consumer hardware.
    
\item {\bf Code of ethics}
    \item[] Question: Does the research conducted in the paper conform, in every respect, with the NeurIPS Code of Ethics \url{https://neurips.cc/public/EthicsGuidelines}?
    \item[] Answer: \answerYes{} % Replace by \answerYes{}, \answerNo{}, or \answerNA{}.
    \item[] Justification: No human subjects. Civilization VI is a commercial game properly cited; CivBench distributes no game assets. MIT license.

\item {\bf Broader impacts}
    \item[] Question: Does the paper discuss both potential positive societal impacts and negative societal impacts of the work performed?
    \item[] Answer: \answerYes{} % Replace by \answerYes{}, \answerNo{}, or \answerNA{}.
    \item[] Justification: CivBench accelerates evaluation of autonomous agents in complex multi-step environments. Potential risks include use of the benchmark to select or optimise agents for long-horizon strategic tasks, including adversarial settings. We mitigate this by releasing the environment and evaluation pipeline openly, enabling broad scrutiny and comparative evaluation.
    
\item {\bf Safeguards}
    \item[] Question: Does the paper describe safeguards that have been put in place for responsible release of data or models that have a high risk for misuse (e.g., pre-trained language models, image generators, or scraped datasets)?
    \item[] Answer: \answerNA{} % Replace by \answerYes{}, \answerNo{}, or \answerNA{}.
    \item[] Justification: No generative model is released. Data consists of tool call logs and diary text, which is original agent output not derived from scraping personal data.

\item {\bf Licenses for existing assets}
    \item[] Question: Are the creators or original owners of assets (e.g., code, data, models), used in the paper, properly credited and are the license and terms of use explicitly mentioned and properly respected?
    \item[] Answer: \answerYes{} % Replace by \answerYes{}, \answerNo{}, or \answerNA{}.
    \item[] Justification: Civilization VI: commercial (user must own a copy); CivBench MCP server: MIT; Inspect evaluation framework: Apache 2.0; game log data: CC-BY 4.0.

\item {\bf New assets}
    \item[] Question: Are new assets introduced in the paper well documented and is the documentation provided alongside the assets?
    \item[] Answer: \answerYes{} % Replace by \answerYes{}, \answerNo{}, or \answerNA{}.
    \item[] Justification: MCP server (MIT); game log dataset for admissible games (CC-BY 4.0); scenario save files (for research use only; no game assets redistributed).

\item {\bf Crowdsourcing and research with human subjects}
    \item[] Question: For crowdsourcing experiments and research with human subjects, does the paper include the full text of instructions given to participants and screenshots, if applicable, as well as details about compensation (if any)? 
    \item[] Answer: \answerNA{} % Replace by \answerYes{}, \answerNo{}, or \answerNA{}.
    \item[] Justification: No human subjects; no crowdsourcing.

\item {\bf Institutional review board (IRB) approvals or equivalent for research with human subjects}
    \item[] Question: Does the paper describe potential risks incurred by study participants, whether such risks were disclosed to the subjects, and whether Institutional Review Board (IRB) approvals (or an equivalent approval/review based on the requirements of your country or institution) were obtained?
    \item[] Answer: \answerNA{} % Replace by \answerYes{}, \answerNo{}, or \answerNA{}.
    \item[] Justification: No human subjects; no crowdsourcing.

\item {\bf Declaration of LLM usage}
    \item[] Question: Does the paper describe the usage of LLMs if it is an important, original, or non-standard component of the core methods in this research? Note that if the LLM is used only for writing, editing, or formatting purposes and does \emph{not} impact the core methodology, scientific rigor, or originality of the research, declaration is not required.
    %this research? 
    \item[] Answer: \answerYes{} % Replace by \answerYes{}, \answerNo{}, or \answerNA{}.
    \item[] Justification: LLMs are both the subject of evaluation and used instrumentally in the analysis pipeline (Claude Haiku 4.5 extracts and labels RAG commitments; Section 6.2 describes this usage fully). Both uses are declared.

\end{enumerate}

\end{document}